\documentclass[10pt,twocolumn,letterpaper]{article}

\usepackage{cvpr}              %

\usepackage[dvipsnames]{xcolor}

\newcommand{\myparagraph}[1]{\vspace{2pt}\noindent{\textbf{#1}}}
\newcommand{\ahm}[0]{aHM}
\newcommand{\acc}[0]{aACC}
\newcommand{\method}[0]{OrCo\xspace}

\usepackage{wasysym}

\usepackage{tabularray}
\usepackage{multirow}
\usepackage{siunitx}
\usepackage{booktabs}
\usepackage[utf8]{inputenc}
\usepackage{arydshln}

\definecolor{cvprblue}{rgb}{0.21,0.49,0.74}
\usepackage[pagebackref,breaklinks,colorlinks,citecolor=cvprblue]{hyperref}

\def\paperID{} %
\def\confName{CLVision 2024}
\def\confYear{2024}

\title{OrCo: Towards Better Generalization via Orthogonality and Contrast for Few-Shot Class-Incremental Learning}

\author{%
Noor Ahmed\textsuperscript{*} \hspace*{.75in} Anna Kukleva\textsuperscript{*} \hspace*{.75in} Bernt Schiele \\[1ex]
\texttt{\{noahmed, akukleva, schiele\}@mpi-inf.mpg.de} \\[1ex] %
Max Planck Institute for Informatics, 
Saarland Informatics Campus}

\begin{document}
\maketitle

\newcommand{\review}[1]{\textcolor[rgb]{0.0, 0.7,0.0}{\textbf{Review: #1}}}

\begin{abstract} 
Few-Shot Class-Incremental Learning (FSCIL) introduces a paradigm in which the problem space expands with limited data. FSCIL methods inherently face the challenge of catastrophic forgetting as data arrives incrementally, making models susceptible to overwriting previously acquired knowledge. 
Moreover, given the scarcity of labeled samples available at any given time, models may be prone to overfitting and find it challenging to strike a balance between extensive pretraining and the limited incremental data.
To address these challenges, we propose the \method framework built on two core principles: features' orthogonality 
in the representation space, and contrastive learning. 
In particular, we improve the generalization of the embedding space by employing a combination of supervised and self-supervised contrastive losses during the pretraining phase. Additionally, we introduce \method loss to address challenges arising from data limitations during incremental sessions.
Through feature space perturbations and orthogonality between classes, the \method loss maximizes margins and reserves space for the following incremental data. This, in turn, ensures the accommodation of incoming classes in the feature space without compromising previously acquired knowledge. Our experimental results showcase state-of-the-art performance across three benchmark datasets, including mini-ImageNet, CIFAR100, and CUB datasets. Code is available at: \url{https://github.com/noorahmedds/OrCo}.
\end{abstract}
 
\let\thefootnote\relax\footnotetext{\textsuperscript{*} Equal Contribution}
\section{Introduction}
\label{sec:intro}

\begin{figure}
\centering
\includegraphics[width=0.98\columnwidth]{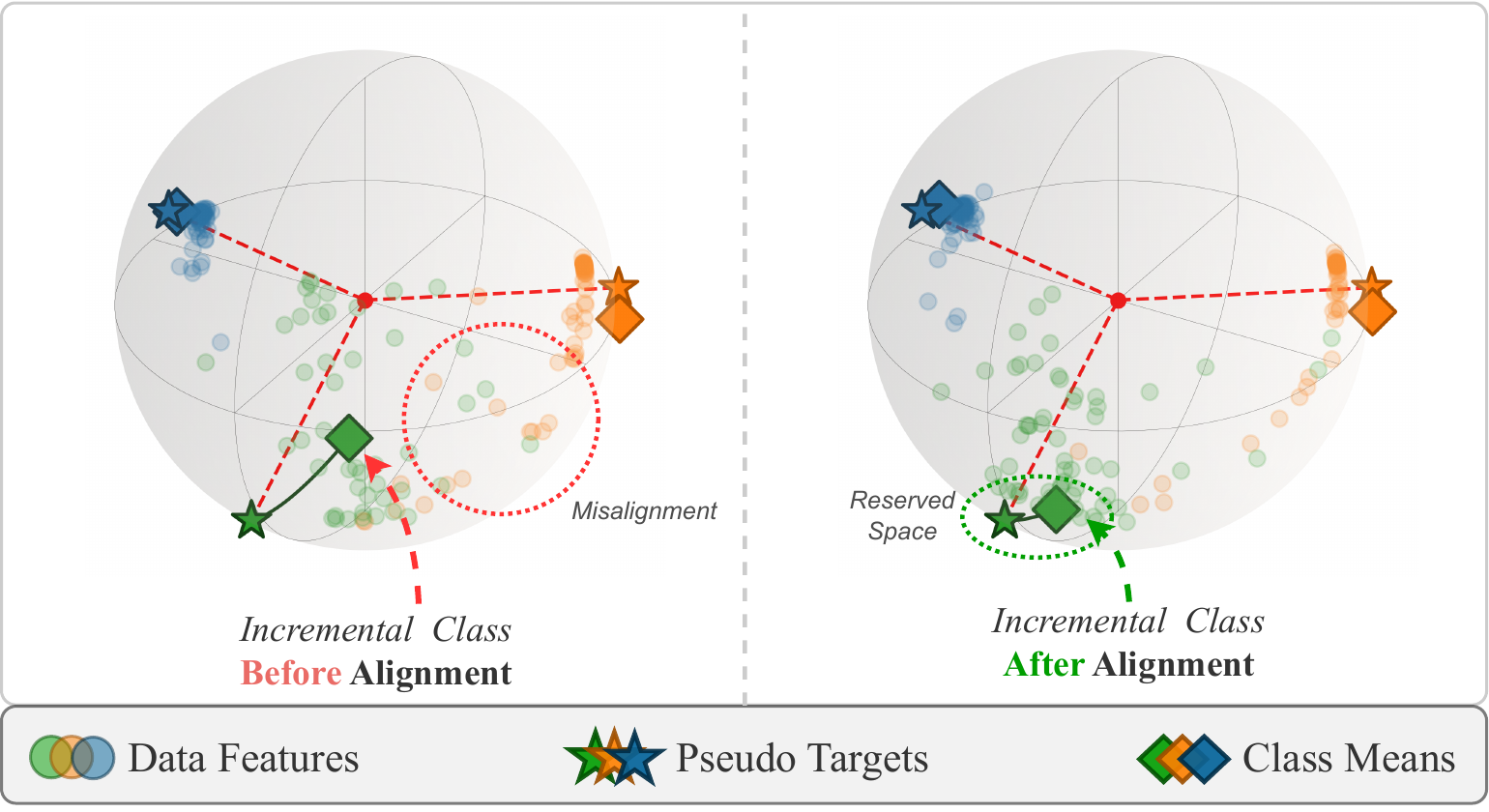}  
\caption{\textbf{PCA analysis on feature space before and after alignment.} \underline{Left}: Before aligning incremental classes to orthogonal pseudo-targets. \underline{Right}: After aligning incremental classes to assigned targets using \textbf{OrCo} loss. Our loss effectively reduces misalignment. Additionally, it enhances generalization for incoming classes by explicitly reserving space.
}
\label{fig:teaser}
\end{figure}

Real-world applications frequently encounter various challenges when acquiring data incrementally, with new information arriving in continuous portions. This scenario is commonly referred to as Class Incremental Learning (CIL)~\cite{ICARL2017, wu2019large, hinton2015distilling, li2017learning, castro2018end, hou2019learning, belouadah2019il2m, zhu2021prototype, zhao2020maintaining, schwarz2018progress, liu2018rotate, chaudhry2018riemannian}. Within CIL, the foremost challenge lies in preventing catastrophic forgetting~\cite{mccloskey1989catastrophic, goodfellow2013empirical, kirkpatrick2017overcoming}, where previously learned concepts are susceptible to being overwritten by the latest updates. \
However, in a Few-Shot Class-Incremental Learning (FSCIL) scenario~\cite{Tao2020, Zhang2021, Anna, zhao2023few, kim2022warping, zhao2021mgsvf, cheraghian2021semantic, Zhou2022, Yang2023Neural, hersche2022cfscil, mazumder2021few, hersche2022cfscil}, characterized by the introduction of new information with only a few labeled samples, two additional challenges emerge: overfitting and intransigence \cite{snell2017prototypical, chen2019closer}. Overfitting arises as the model may memorize scarce input data and lose its generalization ability. On the other hand, intransigence involves maintaining a delicate balance, preserving knowledge from abundant existing classes while remaining adaptive enough to learn new tasks from a highly limited dataset. 
Advances in dealing with catastrophic forgetting, overfitting and intransigence are important steps toward improving the practical value of these methods.

Catastrophic forgetting is commonly tackled in CIL methods~\cite{ICARL2017, hou2019learning, kemker2018fearnet}, which assume ample labeled training data. 
However, standard CIL methods 
struggle in scenarios with limited labeled data, such as FSCIL~\cite{Tao2020}.
To address the three challenges posed by FSCIL, recent approaches~\cite{zhao2023few, Zhang2021} focus primarily on regularizing the feature space during incremental sessions, mitigating the risk of overfitting. These methods rely on a frozen backbone pretrained with standard cross-entropy on a substantial amount of data from the base session. However, we argue that achieving high performance on the pretraining dataset may not necessarily result in optimal generalization in subsequent incremental sessions with limited data. Therefore, in the first phase, we propose enhancing feature space generalization through contrastive learning, leveraging data from the base session.

In this work, we introduce the \method framework, a novel approach built on two fundamental pillars:  features' mutual \underline{or}thogonality on the representation hypersphere and \underline{co}ntrastive learning.
During the first phase, we leverage supervised~\cite{khosla2020supervised} and self-supervised contrastive learning~\cite{pmlr-v9-gutmann10a,chen2020simple,oord2018representation} for pretraining the model. The interplay between these two learning paradigms 
enables the model to capture 
various types of semantic information that is particularly beneficial for the novel classes with limited data~\cite{chen2022perfectly, islam2021broad, Fessss}, implicitly addressing the \textit{intransigence} issue. 
After the pretraining, we generate and fix 
mutually orthogonal random vectors, further referred to as pseudo-targets. In the second phase, we aim to align the fixed pretrained backbone to the pseudo-targets using abundant base data. The learning objective during this phase is our \method loss, which consists of three integral components: perturbed supervised contrastive loss (PSCL), loss term that enforces 
orthogonality
of features in the embedding space, and standard cross-entropy loss. Notably, our PSCL leverages generated pseudo-targets to maximize margins between the classes and to preserve space for incremental data, {enhancing orthogonality through contrastive learning} (see figure~\ref{fig:teaser}).
The third phase of our framework, which we apply in each subsequent incremental session, similarly aims to align the model with the pseudo-targets, but using only few-shot data from the incremental sessions. During the third phase, our PSCL addresses limited data challenges, mitigating the \textit{overfitting} problem to the current incremental session and \textit{catastrophic forgetting} of the previous sessions through margin maximization.

We summarize the contributions of this work as follows:
\begin{itemize}
    \item  We introduce the novel \method framework designed to tackle FSCIL, that is built on orthogonality and contrastive learning principles throughout both pretraining and incremental sessions.
    \item Our perturbed supervised contrastive loss introduces perturbations of orthogonal, data-independent vectors in the representation space. This approach induces increased margins between classes, enhancing generalization.
    \item We showcase robust performance on three datasets, outperforming previous state-of-the-art methods. Furthermore, we perform a thorough analysis to evaluate the importance of each component.
\end{itemize}

\begin{figure*}[t]
\includegraphics[width=\textwidth]{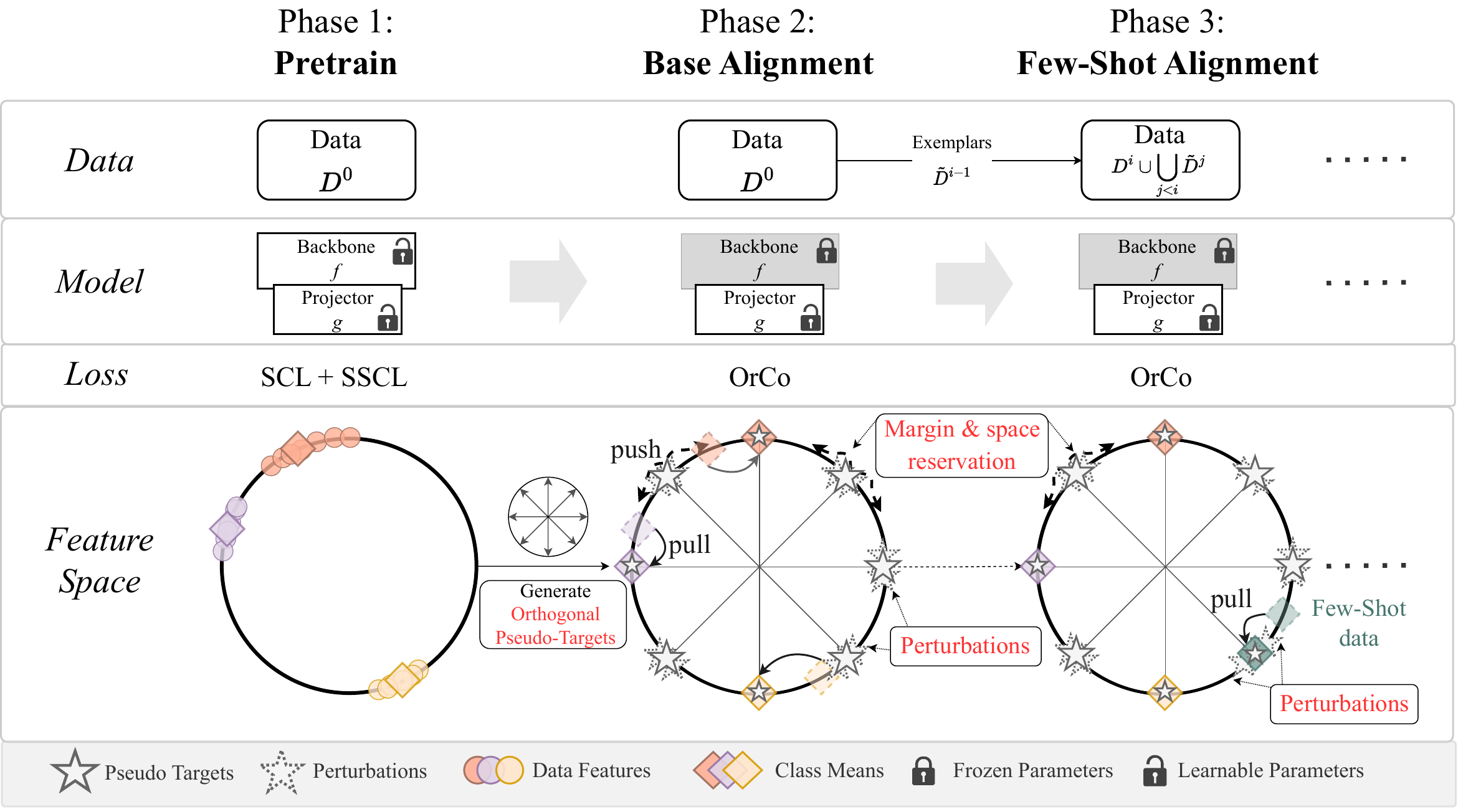} 
\caption{\textbf{Overview of \method framework.} Our \method framework is a three-phase approach for FSCIL. 
\underline{Phase 1 (Pretrain)}:  We pretrain both backbone and projection head with SCL and SSCL on base dataset $D^0$. Before the next phase, we generate mutually orthogonal pseudo-targets. 
\underline{Phase 2 (Base Alignment)}: 
We aim to align the base dataset $D^0$ to the pseudo-targets through our \method loss. This involves pulling class features towards the nearest pseudo-targets and pushing forces based on perturbations around unassigned pseudo-targets (grey stars without assigned colored class means) to increase the margin and preserve space for incoming classes. 
\underline{Phase 3 (Few-Shot Alignment)}: Phase 3, employed in each subsequent incremental session, is similar to Phase 2 and assigns pseudo-targets to incremental class means with further alignment using our \method loss.  
}
\label{fig:main}
\end{figure*}

\section{Related Work}
\myparagraph{Few-Shot Learning (FSL).}
In FSL, a model is trained on scarce data with just a few samples per class. Current literature can be divided into two predominant categories: optimisation-based~\cite{finn2017model, nichol2018firstorder, qi2018low} and metric-based methods~\cite{vinyals2016matching, snell2017prototypical, qi2018low}. 
 Optimization-based approaches, such as MAML~\cite{finn2017model} and Reptile~\cite{nichol2018firstorder}, find optimal parameters that can generalize quickly on other sets when subjected to fine-tuning. 
Conversely, metric-based methods utilize a pretrained model and compare support and query instances using similarity metrics. 
For example, Prototypical Networks~\cite{snell2017prototypical} learns a metric space where the distance to class prototypes determine classification. And
imprinting weights method~\cite{qi2018low} shows improved performance by using class means as strong initialisation for an evolving classifier and integrates principles from both categories. 

\myparagraph{Class-Incremental Learning (CIL).}
In the domain of CIL, a sequence of novel concepts must be learned without forgetting previously acquired knowledge. Recent works can be coarsely categorized in 3 groups. Foremost, there are knowledge distillation schemes \cite{ICARL2017, wu2019large, hinton2015distilling, li2017learning}, which retain model behaviour across the adaptation process to avoid forgetting. Then, data-replay methods \cite{castro2018end, hou2019learning, belouadah2019il2m, zhu2021prototype, zhao2020maintaining} show strong resistance to catastrophic forgetting by storing old class exemplars. Lastly, weight consolidation methods \cite{kirkpatrick2017overcoming, schwarz2018progress, liu2018rotate, chaudhry2018riemannian} identify important weights and moderate training regimen. 

\myparagraph{Few-Shot Class-Incremental Learning (FSCIL).} FSCIL paradigm demands rapid adaptation to novel classes with limited data. 
The methods can be divided into the following categories~\cite{SurveyTian2023}: geometry preservation methods \cite{Tao2020, Zhang2021}, replay or distillation strategies \cite{Anna, zhao2023few}, 
metric learning methods \cite{kim2022warping, zhao2021mgsvf, cheraghian2021semantic, Zhou2022} and meta-learning \cite{Yang2023Neural, hersche2022cfscil, mazumder2021few}, highlighting the breadth of methodologies in FSCIL. 
Parallel to our methodology, 
metric learning methods
utilize tricks in the feature space, showcasing diverse approaches for accommodating incremental classes.
FACT~\cite{Zhou2022} creates virtual prototypes to reserve space and scale the model for incoming classes.  %
NC-FSCIL \cite{Yang2023Neural}, aligns class features 
with the classifier prototypes, which are formed as a simplex equiangular tight frame, using dot-regression loss.
C-FSCIL \cite{hersche2022cfscil} aligns class prototypes quasi-orthogonally to negate interference between classes. In contrast,
our approach stands out for its use of contrastive learning with data agnostic pseudo-targets and margin maximization through perturbations in the embedding space improving generalization in incremental sessions.  %

\section{\method Framework}
We begin with necessary preliminaries in section~\ref{subsec:preliminaries}, followed by the description of our \method framework in section~\ref{subsec:unicon} and \method loss in section~\ref{subsec:unicon_loss}.

\subsection{Preliminaries}
\label{subsec:preliminaries}
\myparagraph{FSCIL Setting.} FSCIL consists of multiple incremental sessions. An initial 0-th session is often reserved to learn a generalisable representation on an abundant base dataset. This is followed by multiple few-shot incremental sessions with limited data. To formalise, an M-Session
N-Way and K-Shot FSCIL task consists of $D_{seq}=\{D^0, D^1,...,D^{M}\}$. These are all the datasets written in sequence where $D^i=\{(x_i, y_i)\}^{|D^i|}_{i = 1}$ is the dataset for the $i$-th session.
The 0-th session dataset $D^0$, also referred to as base dataset, consists of $C^0$ classes, each with a large number of samples. The training set for each following few-shot incremental session ($i>0$)
has $N$ classes. Each of these classes has $K$ samples, typically ranging from 1 to 5 samples per class.
Taking the $i$-th session as an example, the model’s performance is assessed on validation sets from the current ($i$-th) and all previously encountered datasets ($<i$). The entire FSCIL task comprises a total of $C$ classes. 
In our \method framework, we use base dataset $D^0$ for pretraining the model during phase 1 and for base alignment during phase 2.

\myparagraph{Target Generation.}
We employ a Target Generation loss, similarly as in \cite{TSC}, to generate mutually orthogonal vectors across the representation hypersphere with a dimensionality of $d$. First, we define a set of random vectors $T=\{t_i\}$ where $\{t_i\} \in \mathbb{R}^{d}$.
The optimization of the following loss with respect to these random vectors maximizes the angle between any pair of vectors $t_i, t_j \in T$, thereby ensuring their mutual orthogonality: 

\begin{equation}
\mathcal{L}_{TG}(T) = \frac{1}{|T|}\sum^{|T|}_{i=1} \text{log} \sum^{|T|}_{j=1} e^{t_i \cdot t_j / \tau_o}
\label{eq:lsim}
\end{equation}

where $\tau_o$ is the temperature parameter. These optimized vectors, which we further refer to as pseudo-targets, remain fixed throughout our training process.
\\\\
\myparagraph{Contrastive Loss.}
The objective of contrastive representation learning is to create an embedding space where similar sample pairs are in close proximity, while dissimilar pairs are distant. 
In this work, we adopt the InfoNCE loss \cite{khosla2020supervised, oord2018representation} as our contrastive objective. With positive set $P_i$ and negative set $N_i$ defined for each data sample $z_i$, called anchor, this loss aims to bring any $z_j \in P_i$ closer to its anchor $z_i$ and push any $z_k \in N_i$ further away from the anchor $z_i$:
\begin{equation}
    \mathcal{L}_{CL}(i; \theta) = \frac{-1}{|P_i|} \sum_{z_j \in P_i} \text{log} \frac{\text{exp}(z_i\cdot z_j / \tau)}{\sum\limits_{z_k \in N_i} \text{exp}(z_i \cdot z_k / \tau) } ,
\label{eq:supconloss}
\end{equation}
where $\tau$ is the temperature parameter.
In the classical self-supervised contrastive learning (SSCL) scenario, where labels for individual instances are unavailable, the positive set comprises augmentations of the anchor, and all other instances are treated as part of the negative set~\cite{oord2018representation}. In contrast, for the supervised contrastive loss (SCL), the positive set includes all instances from the same class as the anchor, while the negative set encompasses instances from all other classes \cite{khosla2020supervised}. To enhance clarity, we denote the supervised contrastive loss as $\mathcal{L}_{SCL}$ and self-supervised contrastive loss as $\mathcal{L}_{SSCL}$. 
We consider SCL and SSCL as the cornerstone guiding our work due to their discriminative nature, robustness, and extendability. {We leave formal definition of the losses for the supplement.} 

\subsection{\method Framework}
\label{subsec:unicon}

\myparagraph{Overview.} Our \method framework (see figure~\ref{fig:main}) for FSCIL  begins with a pretraining of the model in the first phase, focusing on learning representations, which are transferable to the new tasks.  To achieve this, we leverage both supervised and self-supervised contrastive losses \cite{chen2022perfectly, islam2021broad}.
Before the second phase, we generate a set of mutually orthogonal vectors, which we term as pseudo-targets.
In the subsequent second phase, referred to as base alignment in figure~\ref{fig:main}, we allocate pseudo-targets to class means and ensure alignment through our \method loss, using abundant base data $D^0$. The third phase, implemented in each subsequent incremental session, similarly focuses on assigning incoming but few-shot data to unassigned pseudo-targets, followed by alignment through our \method loss.
Our \method loss comprises three key components: cross-entropy, orthogonality loss, and our novel perturbed supervised contrastive loss (PSCL). The cross-entropy loss aligns incremental data with assigned fixed orthogonal pseudo-targets, the orthogonality loss enforces a geometric constraint on the entire feature space to mimic the pseudo-targets distribution,
and our PSCL enhances crucial robustness for FSCIL tasks through margin maximization and space reservation, 
leveraging mutual orthogonality of pseudo-targets.

\myparagraph{Phase 1: Pretrain.}
In the first pretraining phase, we learn an encoder that accumulates knowledge and generates distinctive features. Using a combination of supervised contrastive loss (SCL) and self-supervised contrastive loss (SSCL), we enhance feature separation within classes, improving model transferability to incremental sessions~\cite{chen2022perfectly, islam2021broad}. To this end, we train the model encoder $f$ and MLP projection head $g$ using base data $D^0$, mapping input images to $\mathcal{R}^{d}$ feature space. The pretraining loss is then defined as:
\begin{equation}
    \mathcal{L}_{pretrain}(D^0; f,g) = (1-\alpha)*\mathcal{L}_{SCL} + \alpha * \mathcal{L}_{SSCL},
\label{eq:supconloss}
\end{equation}
where $\alpha$ controls the contribution of each contrastive loss. 

\myparagraph{Pseudo-targets.}
During the first phase, we do not employ any explicit class vectors that can be used for linear classification. 
Therefore, we generate data-independent mutually orthogonal
pseudo-targets $T = \{t_j\} \in \mathcal{R}^{d}$ on the hypersphere by optimizing loss shown in equation~\ref{eq:lsim}, where $|T| >= C$. 
Further, these pseudo-targets are fixed and assigned to classes,
which, in turn, maximize margins between the classes and improve generalization.

\myparagraph{Phase 2: Base Alignment. }
In addition to the pretraining phase, we introduce the second phase based on the base dataset $D^0$. This phase initiates alignment between the projection head $g$ and the set of generated pseudo-targets $T$.
More specifically, we create class means by averaging features with the same labels. Then, we employ a one-to-one matching approach, utilizing the Hungarian algorithm~\cite{kuhn1955hungarian}, to assign class means with the most fitting set of pseudo-targets $T^0$, where $|T^0| = |C^0|$. 
Note that each class $y_j \in C^0$ is then associated with the respective pseudo-target $t_j^0 \in T^0$ and we denote the remaining unassigned pseudo-targets as $T^0_u = T \setminus T^0$. 
Further, we use pseudo-targets $T^0$ as base class representations for classification. 
Despite the optimal initial assignment, we enhance the alignment of the projection head $g$ and the respective pseudo-targets $T^0$ through the optimization of our \method loss. 
 Further insights into the specifics and motivation behind our \method loss, we elaborate on in  section~\ref{subsec:unicon_loss}.

\myparagraph{Phase 3: Few-Shot Alignment.} 
In the third phase of our framework, applied in each subsequent incremental session, our goal remains to align incoming data with the pseudo-targets.
The following sessions introduce few shot incremental data $D^i$ for the $i$-th session. Building on previous methods~\cite{castro2018end, wu2019large, Anna, cheraghian2021semantic}, we maintain some exemplars from previously seen classes, constituting a joint set $D^{i}_{joint} = \{D^i \cup \{\bigcup_{j=0}^{i-1}\tilde{D}^{j}\}\}$, where $\tilde{D}^{j}$ denotes saved exemplars from earlier sessions.
Keeping random exemplars from previous sessions serves to mitigate both overfitting and catastrophic forgetting issues. 
Similarly to the second phase, we assign pseudo-targets to incremental class means. To be specific, we determine the optimal assignment between $T_u^{i-1}$ and the current class means, resulting in the optimal assignment set of pseudo-targets $T^i$. Respectively, the unassigned set of pseudo-targets becomes $T^i_u = T_u^{i-1} \setminus T^i$. 
During incremental session $i$, we optimize our \method loss given data $D^{i}_{joint}$, pseudo-targets $T=\{T^i\}_0^i \cup T_u^i$ and the assignment between $T=\{T^i\}_0^i$ and the respective classes.

\begin{figure}[!t]
\begin{center}
\includegraphics[width=0.98\columnwidth]{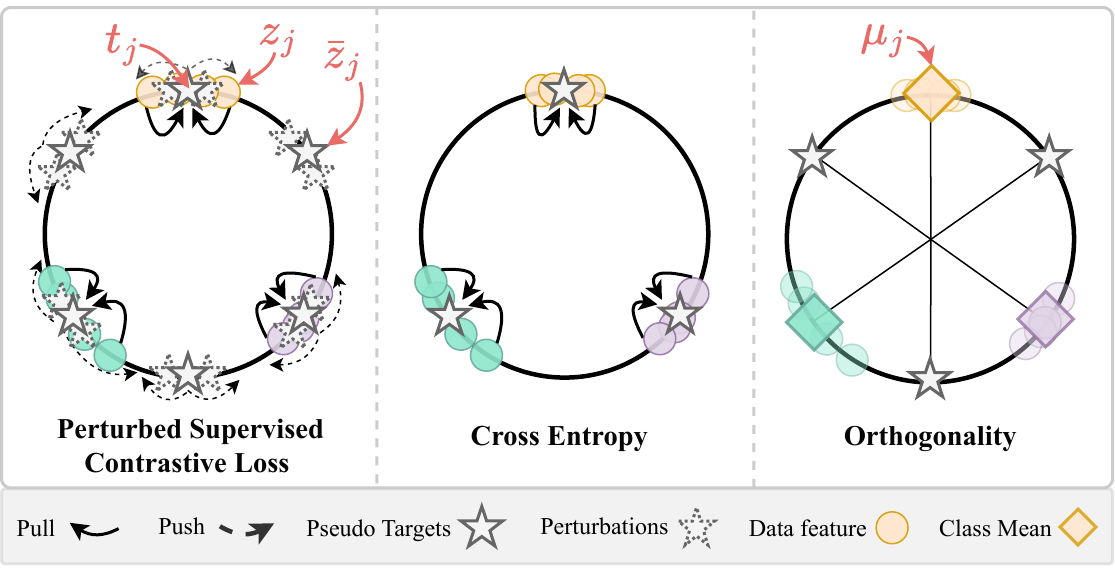} 
\end{center}
\caption{\textbf{\method loss} consists of three components: our proposed perturbed supervised contrastive loss (PSCL), cross-entropy loss (CE), and orthogonality loss (ORTH). $z_j$ denotes the real data anchor point for a contrastive loss, $\bar{z}_j$ denotes the 
unassigned pseudo-target anchor, and $t_j$ denotes an additional positive sample for yellow class in the form of an assigned pseudo-target. 
$\mu_j$ represents the within-batch mean features.}
\label{fig:unicon_loss}
\end{figure}

\subsection{\method Loss}
\label{subsec:unicon_loss}
During the second and the third phases, 
we optimize the parameters of the projection head $g$ with our \method loss. This loss comprises three integral components: our novel perturbed supervised contrastive loss (PSCL), cross-entropy loss (CE) and orthogonality loss (ORTH), see figure~\ref{fig:unicon_loss}. The aim of optimizing the \method loss is to align classes with their assigned pseudo-targets, simultaneously maximizing the margins between classes. This, in turn, enhances overall generalization performance. 

To maximize the margins in the representation space, we introduce uniform perturbations of the pseudo-targets $T$, resulting in perturbed pseudo-targets $\tilde{T} = \{\tilde{t}_j\}_0^{|T|}$ defined as 
\begin{equation}
  \tilde{t}_j = t_j + \mathcal{U}(-\lambda,\lambda),
\label{eq:tpert}
\end{equation}
where $\mathcal{U}$ stands for uniform distribution and $\lambda$ defines sampling boundaries.
To utilize the introduced perturbations, we redefine positive $P_j$ and negative $N_j$ sets for the contrastive loss for the anchor $z_j$ in equation~\ref{eq:supconloss}. Note that during incremental session $i$ the positive set $P_j^i$ in the standard SCL contains all $z_k \in D^i_{joint}$ such that the label $y_k$ is equal to anchor label $y_j$, e.g. in figure~\ref{fig:unicon_loss}, all yellow circles belong to the positive set for yellow anchor $z_j$. And the negative set consists of remaining samples $N_j^i = D^i_{joint} \setminus P_j^i$, in figure~\ref{fig:unicon_loss}, the negative set is composed of all other colors.

To adapt standard SCL to PSCL, we expand the definition of the positive set. The anchor $z_j$, with its assigned pseudo-target $t_j \in T$, becomes an additional positive pair, see figure~\ref{fig:unicon_loss}. Furthermore, considering the previously defined pseudo-target perturbations, we incorporate them into the positive set, resulting in $\tilde{P}_j^i = P_j^i \bigcup t_j \bigcup \tilde{t}_j$. In figure~\ref{fig:unicon_loss}, the positive set for yellow anchor $z_j$ contains all yellow circles and additionally the pseudo-target $t_j$ with its perturbations.  
This extension of the positive set introduces additional pushing forces for incremental classes and, therefore, enables the maximization of margins between classes. We show that this approach proves to be especially advantageous in scenarios with limited samples, as it mimics augmentations in the feature space.

On the other hand, we expand the anchor definition. In standard SCL, each anchor $z_j$ belongs to the set of real training data $D^i_{joint}$, e.g. in figure~\ref{fig:unicon_loss}, anchors for standard SCL are only circles.
However, we propose to use anchors from both real data and unassigned pseudo-targets (circles and unassigned stars in figure~\ref{fig:unicon_loss}), 
specifically $z_j \in D^i_{joint}$ and $\bar{z}_j \in T^i_u$. 
The positive set for the anchor $\bar{z}_j \in T^i_u$ (unassigned pseudo-target) contains only corresponding perturbed pseudo-targets $\tilde{P}_j^i = \{\tilde{t}_j\}$ (dashed stars around $\bar{z}_j$ in figure~\ref{fig:unicon_loss}), while the negative set $\tilde{N}_j^i = \{D^i_{joint} \bigcup \tilde{T}^i_u\} \setminus \{\tilde{t}_j\}$ includes all real data and other perturbed pseudo-targets. This approach ensures that each unassigned pseudo-target pushes all other classes away, thereby promoting space preservation for the following incremental sessions.

To complement PSCL, we employ cross-entropy loss for sample $z_j$ that pulls class features to their assigned targets during the few-shot incremental session i:
\begin{equation}
    \mathcal{L}_{CE}(z_j) = -\sum_{c \in \{C^I\}_1^i} y_{c} \text{log} \frac{\text{exp}(z_jt_c^T)}{\sum_{k \in \{C^I\}_1^i} \text{exp}(z_kt_c^T) }. 
\label{eq:celoss}
\end{equation}

We further employ the orthogonality loss ($\mathcal{L}_{ORTH}$) defined similarly as in equation \ref{eq:lsim}. 
It differs, however, in that it uses the mean class features $\mu_j$ (see figure \ref{fig:unicon_loss}) as input and enforces an intrinsic geometric constraint on the feature landscape. See section \ref{sec:detail_orth_loss} in supplementary for details.

Finally, our \method loss is a combination of the three losses introduced above: 
\begin{equation}
    \mathcal{L}_{\method} = \mathcal{L}_{PSCL} + \mathcal{L}_{CE} + \mathcal{L}_{ORTH}.
\label{eq:uniconloss}
\end{equation}

During testing, a sample is assigned a label based on the nearest assigned pseudo target.
\section{Experimental Results}
In section~\ref{subsec:datasets}, dataset and evaluation protocol are introduced. Then we compare with state-of-the-art methods on 3 popular benchmarks in section~\ref{subsec:sota}. Finally, we validate the effectiveness of each of the components in section~\ref{subsec:ablationns}.

\begin{table*}
\centering

\begin{tblr}{
  row{2} = {c},
  column{11} = {c},
  column{2} = {c},
  cell{1}{1} = {r=2}{},
  cell{1}{2} = {r=2}{},
  cell{1}{3} = {c=8}{c},
  cell{1}{11} = {r=2}{},
  cell{1}{12} = {r=2}{},
  hline{1,12} = {-}{0.1em},
  hline{2} = {3-10}{0.06em},
  hline{3} = {-}{0.06em},
  hline{4,11} = {-}{dotted},
}
Method & Base Acc & Session-wise Harmonic Mean~(\%)~↑ &                &                &                &                &                &                &                & \ahm{}      & $\Delta$\ahm{} \\
                 &                   & 1                                          & 2              & 3              & 4              & 5              & 6              & 7              & 8              &                &                    \\
IW~\cite{qi2018low}               & 83.10             & 49.49                                      & 45.09          & 45.98          & 46.30          & 44.67          & 42.48          & 43.26          & 45.65          & 45.36          & \textbf{+12.76}             \\
FACT~\cite{Zhou2022}             & 75.78             & 27.20                                      & 27.84          & 27.94          & 25.17          & 22.46          & 20.54          & 20.88          & 21.25          & 24.16          & \textbf{+33.96}             \\
CEC~\cite{Zhang2021}              & 72.17             & 31.91                                      & 31.84          & 30.98          & 30.74          & 28.14          & 26.78          & 26.96          & 27.42          & 29.35          & \textbf{+28.78}             \\
C-FSCIL~\cite{hersche2022cfscil}          & 76.60             & 9.74                                       & 20.53          & 28.68          & 31.91          & 34.85          & 35.05          & 37.72          & 37.92          & 29.55          & \textbf{+28.57}             \\
LIMIT~\cite{zhou2022few}            & 73.27             & 40.34                                      & 33.58          & 31.81          & 31.74          & 29.32          & 29.11          & 29.57          & 30.28          & 31.97          & \textbf{+26.15}             \\
LCwoF~\cite{Anna}            & 64.45             & 41.24                                      & 38.96          & 39.08          & 38.67          & 36.75          & 35.47          & 34.71          & 35.02          & 37.49          & \textbf{+20.63}             \\
BiDist~\cite{zhao2023few}           & 74.67             & 42.42                                      & 43.86          & 43.87          & 40.34          & 38.97          & 38.01          & 36.85          & 38.47          & 40.35          & \textbf{+17.77}             \\
NC-FSCIL~\cite{Yang2023Neural}         & \textbf{84.37}             & 62.34                                      & 61.04          & 55.93          & 53.13          & 49.68          & 47.08          & 46.22          & 45.57          & 52.62          & \textbf{+5.50}              \\
OrCo           & 83.30       & \textbf{68.71}                             & \textbf{63.87} & \textbf{60.94} & \textbf{57.98} & \textbf{55.27} & \textbf{52.41} & \textbf{52.68} & \textbf{53.12} & \textbf{58.12} &                    
\end{tblr}
\caption{\textbf{Sota comparison on mini-ImageNet.} 
\ahm{} denotes the average of the harmonic mean across all sessions. 
IW~\cite{qi2018low} is evaluated based on the model learning in our pretrain phase.
Detailed results of the individual sessions are in the supplement.
}
\label{tab:miniImagenetSota}
\end{table*}

\subsection{Datasets and Evaluation}
\label{subsec:datasets}
We conduct evaluation of our \method framework on three FSCIL benchmark datasets: mini-ImageNet \cite{russakovsky2015imagenet}, CIFAR100 \cite{krizhevsky2009learning} and CUB200 \cite{wah2011caltech}. In the setting formalised by \cite{Tao2020} mini-ImageNet and CIFAR100 are organized into 60 base classes and 40 incremental classes structured in a 5-way, 5-shot FSCIL scenario for a total of 8 sessions. CUB200, a dataset for fine-grained bird species classification, contains equal number of base and incremental classes for a total of 200 classes. The dataset is organised as a 10-way, 5-shot FSCIL task and presents a rigorous challenge.

\begin{figure}[t]
    \centering
    \includegraphics[width=\linewidth]{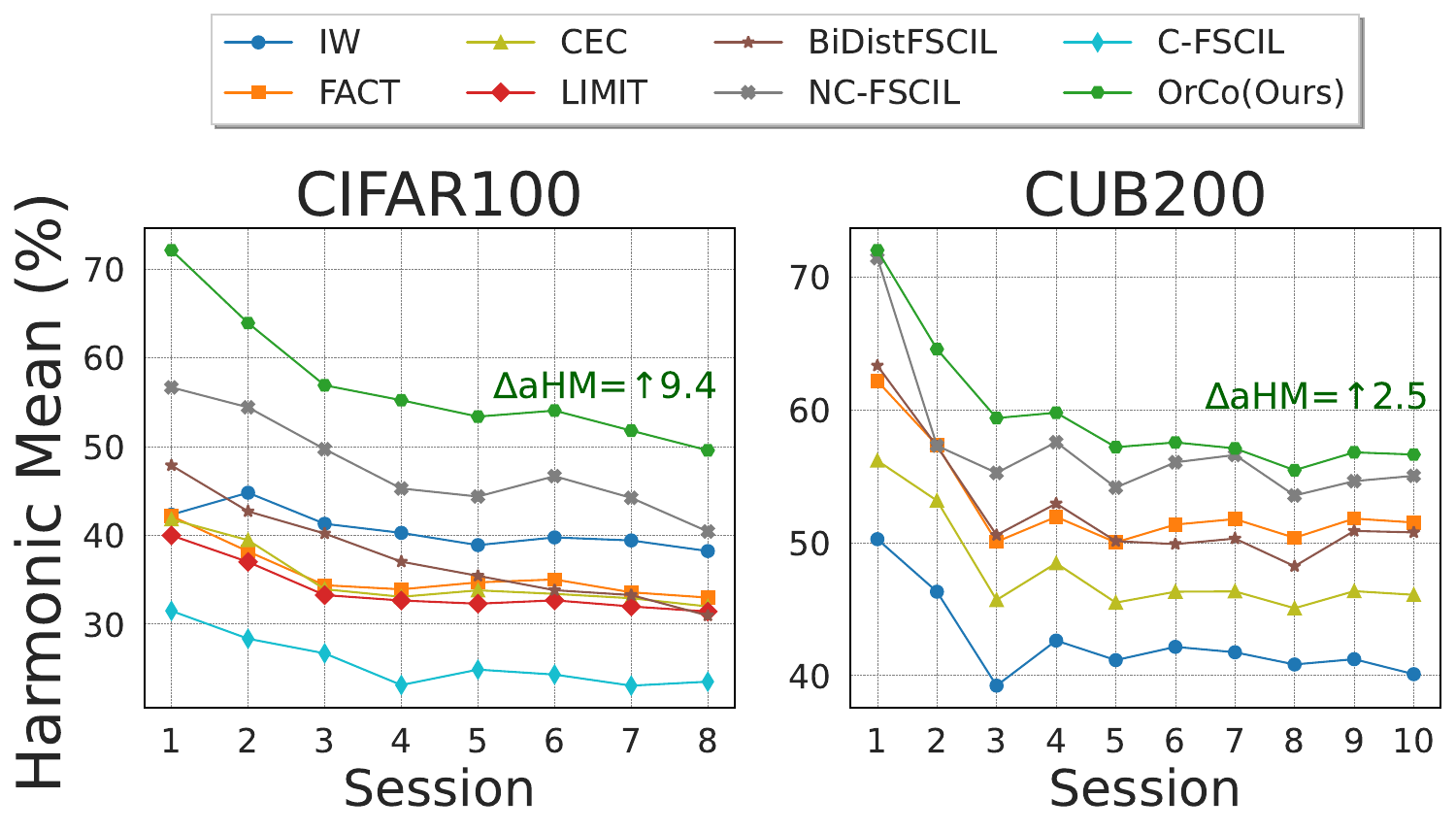}
  \caption{\textbf{Sota comparisons on CIFAR100 and CUB200 datasets.} Performance curves, that measure harmonic mean, of our method comparing to recent sota methods. Left: CIFAR100. Right: CUB200. $\Delta$\ahm{} denotes the average harmonic mean improvement over the runner-up method. 
  }
  \label{fig:sotaplot}
  \vspace{-4mm}
\end{figure}

\myparagraph{Performance measure} Commonly used FSCIL datasets all have a quantity bias towards the base classes. mini-ImageNet and CIFAR100 have both 60\% of the data in the base classes and CUB200 with 50\%. Consequently, standard accuracy measures like Top-1 accuracy will be skewed in favour of the base-classes. For instance, a method which has a base accuracy $A_{base} = 100\%$ on CUB200 and performs weakly on the first incremental session $A_{inc}^1 = 10\%$ would produce a Top-1 average accuracy $A_{cls}^{1} = 91.82\%$. At first glance, this accuracy may not entirely represent inherent biases in a method, though such measures are commonly used to benchmark performance.
To tackle this, harmonic mean has risen as a robust evaluation measure in FSCIL \cite{Anna, Zhou2022, zhao2023few}. In the given scenario, the harmonic mean would penalise the method aggressively resulting in a metric score of $A^{1}_{hm} = 18.18\%$, accurately indicating bias. More concretely, we compute harmonic mean by combining base class accuracy and incremental session accuracy: $A^j_{hm} = (2\times A_{base}\times A^j_{inc})/(A_{base} + A^{j}_{inc})$. 
In addition to this, we propose average harmonic mean (\ahm{}) which is simply averaging the harmonic mean scores from all sessions for a consolidated view.

\myparagraph{Implementation Details} 
Our model is optimised using LARS \cite{you2017large} for the pretraining phase and SGD with momentum for phase 2 and 3. For CUB200 dataset, we skip the pretraining following \cite{Yang2023Neural, Zhang2021, Tao2020} and initialize the model with ImageNet pretrained weights. 
For the second and third phase, we finetune only the projection head.
For the PSCL loss we choose a perturbation magnitude $\lambda = \text{1e-2}$.
We train the projection head for 10 epochs during the second phase and 100 epochs for the third phase. Cosine scheduling is employed with a maximum learning rate set to $0.1$. Augmentations include, random crop, random horizontal flip, random grayscale and a random application of color jitter. Details can be found in the supplement section \ref{sec:impl_details}.

\setlength\extrarowheight{3pt}

\begin{table*}
\centering
\parbox{.76\linewidth}{
\small
\begin{tabular}{ccclllllllll} 
\toprule
\multicolumn{1}{c}{\multirow{2}{*}{PSCL}} & \multicolumn{1}{c}{\multirow{2}{*}{CE}} & \multicolumn{1}{c}{\multirow{2}{*}{ORTH}} & \multicolumn{8}{c}{\textbf{Session-wise Harmonic Mean~(\%)~↑ }}                                                                                                                               & \multicolumn{1}{c}{\multirow{2}{*}{aHM}}  \\ 
\cline{4-11}
\multicolumn{1}{c}{}                      & \multicolumn{1}{c}{}                    & \multicolumn{1}{c}{}                      & \multicolumn{1}{c}{1} & \multicolumn{1}{c}{2} & \multicolumn{1}{c}{3} & \multicolumn{1}{c}{4} & \multicolumn{1}{c}{5} & \multicolumn{1}{c}{6} & \multicolumn{1}{c}{7} & \multicolumn{1}{c}{8} & \multicolumn{1}{c}{}                      \\ 
\midrule
                                          &     \checkmark                                    &                                           & 65.46                 & 56.29                 & 44.12                 & 36.96                 & 26.90                 & 21.11                 & 18.90                 & 16.19                 & 35.74                                     \\ 
\hdashline[1pt/1pt]
                                          &        \checkmark                                 &        \checkmark                                   & 65.30                 & 56.21                 & 43.96                 & 37.30                 & 28.31                 & 22.01                 & 19.66                 & 16.64                 & 36.17                                     \\ 
\hdashline[1pt/1pt]
             \checkmark                             &                                         &                                           & 50.70                 & 45.42                 & 42.68                 & 39.84                 & 38.71                 & 37.94                 & 36.26                 & 35.87                 & 40.93                                     \\ 
\hdashline[1pt/1pt]
               \checkmark                           &                                         &   \checkmark                                        & 52.34                 & 47.24                 & 43.79                 & 41.62                 & 41.15                 & 39.68                 & 38.69                 & 37.34                 & 42.73                                     \\ 
\hdashline[1pt/1pt]
                   \checkmark                       &         \checkmark                                &                                           & 68.04                 & \textbf{63.94}                 & 60.22                 & \textbf{58.00}                 & \textbf{55.44}                 & 51.51                 & 51.88                 & 52.74                 & 57.72                                     \\ 
\hdashline[1pt/1pt]
                      \checkmark                    &               \checkmark                          &         \checkmark                                  & \textbf{68.71}        & 63.87        & \textbf{60.94}        & 57.98        & 55.27        & \textbf{52.41}        & \textbf{52.68}        & \textbf{53.12}        & \textbf{58.12}                            \\
\bottomrule
\end{tabular}
\caption{\textbf{Influence of OrCo loss components.} PSCL denotes perturbed supervised contrastive loss, CE denotes cross-entropy, ORTH denotes orthogonality loss. See figure~\ref{fig:unicon_loss} for visualization of each component. Ablation study on mini-ImageNet. }
\label{tab:lossablation}
}
\hfill
\parbox{.2\linewidth}{
\centering
\small
  \begin{tabular}{
  c c
  }
  \toprule
  Sampling                           &  \multirow{1}{*}{\ahm} \\   
  \midrule
  Rand & 57.23          \\
  Orth & \textbf{58.12} \\
  \bottomrule
\end{tabular}
  \caption{\textbf{Importance of explicit orthogonality loss for pseudo-target generation.}  Rand denotes random sampling from normal distribution. 
  }
  \label{tab:uniformityablation}
}
\vspace{-3mm}
\end{table*}

\subsection{Comparison to state-of-the-art}
\label{subsec:sota}

In this section, we conduct a comparative analysis of our proposed \method with recent state-of-the-art approaches. Table~\ref{tab:miniImagenetSota} presents the results obtained on the mini-ImageNet dataset, while figure~\ref{fig:sotaplot} illustrates the evaluation results on the CUB200 and CIFAR100 datasets. Our method demonstrates superior performance across all three datasets, surpassing previous state-of-the-art methods by a significant margin, particularly achieving improvements of 9.4\% and 5.5\% on CIFAR100 and mini-ImageNet, respectively. Notably, the effectiveness of \method is consistently evident across all incremental sessions.

In addition to reporting results for the standard FSCIL methods, we also present the performance of the Imprinted Weights method (IW)~\cite{qi2018low} based on our model pretrained during Phase 1. The robust performance of this method indicates the efficacy of our pretraining strategy in facilitating effective transferability to downstream tasks, such as incremental few-shot learning sessions, thereby addressing the intransigence problem.
We present a detailed breakdown of each session in section \ref{sec:more_results} of the supplement.

\subsection{Analysis}
\label{subsec:ablationns}
To validate the effectiveness of each component of our framework, in this section, we show an analysis based on the mini-ImageNet dataset. 

\myparagraph{\method loss.}
We assess the efficacy of the components comprising our \method loss in table~\ref{tab:lossablation}. The \method loss consists of three integral components illustrated in figure~\ref{fig:unicon_loss}: the cross-entropy loss ({CE}), the orthogonality loss ({ORTH}), and the perturbed supervised contrastive loss ({PSCL}). 
We observe that CE struggles to generalize on underrepresented incremental classes. On the contrary, PSCL enhances the robust SCL approach with pseudo target perturbations and provides better class separation. PSCL, on its own, shows steady generalization, with only a 14.83\% drop in harmonic mean. CE, however, while starting strong, ultimately becomes biased towards base classes, leading to a significant 49.26\% drop in harmonic mean.
By integrating the dynamic yet fundamentally discerning features of CE with the stability offered by PSCL, a significant enhancement in harmonic mean is achieved.
Moreover, the orthogonality loss (ORTH) 
consistently improves performance (+0.4\%). Its integration into our loss formulation further underscores the significance of orthogonality.

\begin{figure}
\centering
\includegraphics[width=0.8\columnwidth]{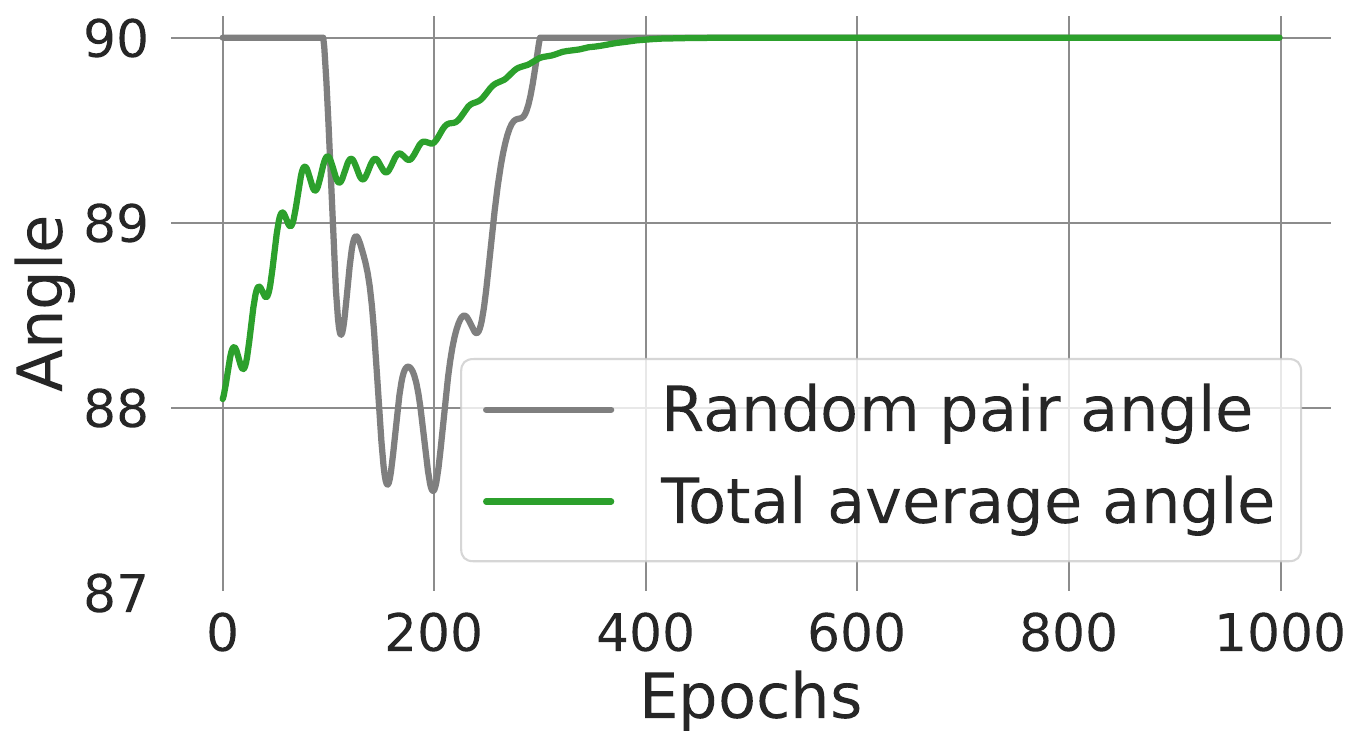} 
\caption{\textbf{Measurement of angle during orthogonality optimization.} The green curve corresponds to the evolution of the average angle between all pairs during the optimization. The gray curve shows measurements of random pairs at each epoch.  
}
\label{fig:evolvetargets}
\end{figure}

\myparagraph{Influence of mutually orthogonal pseudo-targets.} 
We note that independent randomly sampled vectors from a Gaussian distribution $\mathcal{N}(0,1)$ are theoretically orthogonal on the surface of a unit sphere (refer to the supplement for a discussion).  However, in practice, we observe only a near-orthogonal behavior, as illustrated in our training curve for orthogonal pseudo-target generation in figure~\ref{fig:evolvetargets}. To examine the impact of a perfectly orthogonal target space, we assess our method using both randomly sampled targets from a Gaussian distribution and our generated orthogonal targets, as presented in table~\ref{tab:uniformityablation}. We observe a 0.89\% improvement in performance when explicit orthogonality constraints are applied. This finding suggests that an aligned and orthogonal feature space is more effective in addressing data imbalances between base and incremental sessions. Consequently, we incorporate orthogonality as a fundamental principle in our framework, recognizing its significant role in enhancing the overall effectiveness of our model.

\myparagraph{Pseudo-targets perturbations. }
\method relies on perturbations of fixed pseudo-targets to introduce a margin between previously encountered and incoming classes. We compare \method against a variant where the contrastive loss does not receive any pseudo-targets' perturbations (w/o). In contrast to this, our perturbation schemes with sampling $\lambda$, as in equation~\ref{eq:tpert}, from Gaussian ($\mathcal{N}$) and uniform ($\mathcal{U}$) distributions consistently enhance the final session harmonic mean ($HM_8$) by over 30\%, as shown in table~\ref{tab:pertablation}.

For a detailed evaluation of our method, we employ false positive and cosine similarity analyses. By measuring the false positive rate within only incremental classes ($FP_{inc}$), we observe improved separation between the few-shot classes with the perturbed objective.

Subsequently, we calculate the average inter-class cosine similarity ($Sim_{cls}$) for all base and few-shot incremental classes, providing an indication of the spread of each class on the unit sphere. A lower value suggests more compact representations. Notably, we observe values at least 10 times lower for the training with perturbations.
Lastly, we assess the availability of space around unassigned pseudo-targets ($Sim_{cls\rightarrow target}$) by computing the average similarity of all data features with respect to all unassigned targets. A higher average similarity corresponds to smaller margins between the features and the unassigned pseudo-targets. Table~\ref{tab:pertablation} illustrates that perturbations indeed increase the margin around unassigned pseudo-targets. Further discussions can be found in the supplement.

\begin{table}
\small
\setlength{\tabcolsep}{3pt} %
\begin{tabular}{
    c c c c c
}
\toprule
Perturbation                             & $FP_{inc}$~$\downarrow$ & $Sim_{cls}$~$\downarrow$ & $Sim_{cls\rightarrow target}\downarrow$ & $HM_8\uparrow$ \\
\midrule
w/o   & 66.5 & 0.105 & 0.013 & 20.14        \\      
$\mathcal{N}$ & 54.6 & \textbf{0.002} & \textbf{0.006} & 50.23         \\
$\mathcal{U}$ & \textbf{52.5} & 0.011 & \textbf{0.006} & \textbf{53.12} \\
\bottomrule
\end{tabular}
\caption{\textbf{Influence of perturbations in PSCL.} Comparison of our PSCL loss with or without perturbations of pseudo-targets. $\mathcal{N}$, $\mathcal{U}$ denotes Gaussian and Uniform distributions, respectively, from which $\lambda$, as in equation~\ref{eq:tpert}, sampled during training. $FP_{inc}$ refers to the False Positive rate among all incremental classes. $Sim_{cls}$ computes the average pairwise cosine similarity between all class pairs. $Sim_{cls\rightarrow target}$ indicates the pairwise cosine similarity between classes and unassigned target pairs over all sessions. $HM_8$ refers to the 8-th and final session harmonic mean.}
\label{tab:pertablation}
\vspace{-3mm}
\end{table}

\myparagraph{Influence of pretraining. }
To evaluate our pretraining strategy, we compare it against cross entropy (CE) and standard supervised contrastive loss (SCL)~\cite{khosla2020supervised}. As shown in table~\ref{tab:pretrainablation}, the addition of self-supervised contrastive loss (SCL+SSCL) to the pretraining session significantly enhances generalization on unseen data, showcasing improved transfer capabilities, which aligns with previous findings~\cite{chen2022perfectly, islam2021broad}. Additionally, we present the accuracy on the validation set for the base classes $D^0$ immediately after the pretrain phase for each strategy.

While all strategies exhibit close to 85\% accuracy on the base validation set, our approach yields a 0.74\% higher average harmonic mean compared to \textit{SCL} and a notable 2.92\% improvement over \textit{CE}. The significance of this lies in the fact that our frozen backbone network, maintained during incremental sessions, is capable of producing strong and unique features even for unseen classes.

\begin{table}
\centering
\small
\begin{tabular}{
  c c c
}
\toprule
Pretrain Strategy                            & Phase 1:Accuracy & \ahm{} \\    
\midrule
CE & 85.70 & 55.20 \\
SCL & 85.18 & 57.38 \\
SCL + SSCL (Ours) & \textbf{85.95} & \textbf{58.12} \\  
\bottomrule
\end{tabular}
\caption{\textbf{Influence of pretraining.} \ahm{} denotes average harmonic mean. CE is cross-entropy, SCL is supervised countrastive loss, and SSCL is self-supervised contrastive loss. 
}
\label{tab:pretrainablation}
\end{table}

\myparagraph{Frozen parameters.}
Table~\ref{tab:decoupledablation} illustrates how \method effectively addresses catastrophic forgetting by adopting a strategy of freezing the backbone and training only the projection head.
The observed overall performance decay, along with a 2.7\% greater loss of base accuracy across all sessions, demonstrates favorable outcomes for decoupling the learning process after Phase 1.
\begin{table}
\centering
\small
\begin{tblr}{
  column{1,2,3} = {c},
  hline{1,4} = {-}{0.08em},
  hline{2} = {-}{},
}
Fine-tuned params                             & Performance Decay $\downarrow$ & Base Decay $\downarrow$\\    
$f,g$ & 28.94 & 20.52\\
$g$  & \textbf{26.99} & \textbf{17.83}\\  

\end{tblr}
\caption{\textbf{Influence of frozen paramters.} Analysing of catastrophic forgetting when 1) fine-tuning the entire model ($f,g$) and 2) fine-tuning only the projection head ($g$).}
\label{tab:decoupledablation}
\vspace{-3mm}
\end{table}

\section{Conclusion}

This paper introduced \method method to boost the performance of FSCIL by addressing its inherent challenges:
catastrophic forgetting,  overfitting, and intransigence. The \method framework is a novel approach that tackles these issues by leveraging features' mutual orthogonality on the representation hypersphere and contrastive learning. By combining supervised and self-supervised contrastive learning during pretraining, the model captures diverse semantic information crucial for novel classes with limited data, implicitly addressing the intransigence challenge. Employing 
the proposed \method loss during subsequent incremental sessions ensures alignment with the generated fixed pseudo-targets, maximizing margins between classes and preserving space for incremental data. This comprehensive approach not only enhances feature space generalization but also mitigates overfitting and catastrophic forgetting, marking steps toward improving the practical value of incremental learning methods in real-world applications.

{
    \small
    \bibliographystyle{ieeenat_fullname}
    \bibliography{main}
}

\clearpage
\maketitlesupplementary

Within the supplement, we provide additional ablation studies in section~\ref{sec:ablations_supmat}, detailed breakdown tables and confusion matrices in section~\ref{sec:more_results}, an extended discussion on theory of orthogonality in section~\ref{sec:ortho}, formulation of contrastive losses in section~\ref{sec:contrastive_losses} and additional implementation details in section~\ref{sec:impl_details}.

\section{More Ablations}
\label{sec:ablations_supmat}

\myparagraph{What to pull \& what to perturb in OrCo loss.}
Our OrCo loss comprises both pull and push components, influencing the distribution over the hypersphere.
The pull effect is driven by cross-entropy loss (CE), where data features align with their assigned pseudo-targets. As illustrated in 
table ~\ref{tab:pull-perturb}, we show the advantage of aligning data features and pseudo-targets specifically from incremental sessions during the third phase. Introducing pseudo-targets assigned to the base classes to CE loss results in a performance degradation of approximately 1\% in HM$_8$, due to an increased bias towards base classes.
Next, we study the impact of perturbations, which create additional pushing forces, on different subsets of pseudo-targets. Our findings indicate that perturbing both incremental- and base-assigned pseudo-targets consistently hampers performance compared to perturbing only those assigned to incremental classes, resulting in about 9\% improvement in HM$_8$. Higher base accuracy indicates that perturbations of both base- and incremental-assigned pseudo-targets provide more room for prevalent base classes, hindering the learning of novel classes and favouring base-class bias.

\myparagraph{Pseudo-targets assignment strategy.}
In table~\ref{tab:assignablation}, we highlight the crucial role of optimal initial alignment between pseudo-targets and class means. We compare a random assignment strategy to a Hungarian matching algorithm. Hungarian matching allows to find an optimal assignment based on distances between class means and pseudo-targets. We identify two optimal assignment strategies within hungarian matching 1) Reassignment and 2) Greedy Assignment. For the former, class means are reassigned to closest pseudo-targets at the beginning of each session whereas the later, carries forward the assignment from previous sessions.

We find that the random assignment strategy leads to a notable degradation in accuracy, particularly evident after the second phase for the base classes, amounting to approximately 8\%. 
Greedy assignment performed better than reassignment. Despite reassignment being theoretically optimal, in practice we observe a performance drop likely due to noisy few-shot classes appearing geometrically closer to previously assigned pseudo-targets hence causing a shift of previously seen assigned classes and causing misalignment. This can be clearly seen in the loss of generalisation given a base decay of 29.65\% vs 15.62\% for best case.

Overall accuracy is substantially improved, demonstrating the critical contribution of the optimal assignment  approach in addressing forgetting and achieving better alignment.

\begin{table}[t]
\scalebox{1}{ 
\centering
\setlength{\tabcolsep}{1mm}{
\resizebox{0.98\columnwidth}{!}{
\small
\begin{tabular}{c c c c c c c}
\toprule
\multirow{2}{*}{CE} & \multirow{1}{*}{Perturbed} & \multirow{2}{*}{Base Acc} & \multirow{2}{*}{Inc Acc} & \multirow{2}{*}{$HM_8$} & \multirow{2}{*}{\ahm{}} & \multirow{2}{*}{\acc{}}\\
& Pseudo-Targets & & & \\
\midrule
\multirow{2}{*}{\textbf{Inc}} & \textbf{Inc} & 67.60 & \textbf{43.80} & \textbf{53.12} & \textbf{58.12} & 67.14\\
 & Base+Inc & 78.13 & 30.86 & 44.26 & 50.44 & 69.17\\
\hdashline[1pt/1pt]
\multirow{2}{*}{Base+Inc} & Inc & 69.65 & 41.85 & 52.30 & 57.76 & 67.90\\
 & Base+Inc & \textbf{78.90} & 29.53 & 43.00 & 48.77 & \textbf{69.25}\\
\bottomrule
\end{tabular}
}}}
\caption{\textbf{What to pull \& what to perturb.} CE denotes cross-entropy that pulls data features to the pseudo-targets; 
Inc denotes that only assigned to incremental sessions pseudo-targets participate in the CE loss, Base+Inc denotes both base- and incremental-assigned pseudo-targets. The choice of perturbed pseudo-targets can include incremental assigned pseudo-targets with unassigned pseudo-targets (Inc), or all assigned pseudo-targets with unassigned pseudo-targets (Base+Inc).  Base/Inc Acc denotes accuracy from the last $8^{th}$ session. aACC denotes average accuracy over all sessions. Results on mini-ImageNet.
}
\label{tab:pull-perturb}
\end{table}

\begin{table}
\scalebox{1}{ 
\centering
\setlength{\tabcolsep}{1mm}{
\resizebox{0.98\columnwidth}{!}{
\small

\begin{tblr}{
  row{0} = {c},
  row{2} = {c},
  row{3} = {c},
  row{4} = {c},
  hline{1,5} = {-}{0.08em},
  hline{2} = {-}{},
  hline{3} = {-}{dotted},
}
Assignment                         & Base Acc~$\uparrow{}$ & Base Decay~$\downarrow{}$  &  \ahm{}~$\uparrow{}$ & \acc{}~$\uparrow{}$\\
Random  & 75.75 & 20.40 & 54.40 & 59.42          \\
Reassignment & \textbf{83.30} & 29.65 & 55.49 & 62.74 \\
Greedy & \textbf{83.30} & \textbf{15.72} & \textbf{58.12} & \textbf{67.14}
\end{tblr}
}}}
\caption{\textbf{Pseudo-targets assignment strategy.} Comparing our optimal assignment strategy against random assignment of pseudo-targets.}
\label{tab:assignablation}
\end{table}

\myparagraph{Number of exemplars.}
Due to the memory constraints inherent in FSCIL, it is common to utilize a constrained number of exemplars from the previous task. To investigate this, we conducted tests with 0, 1, and 5 exemplars, and the results are presented in table~\ref{tab:exemplarablation}. We note that even with just 1 exemplar, our model achieves a performance improvement of 2.84\% compared to our strong baseline, the IW method.
\begin{table}[t]
\scalebox{1}{ 
\centering
\setlength{\tabcolsep}{1mm}{
\resizebox{0.98\columnwidth}{!}{
\small
\begin{tblr}{
  row{1} = {c},
  cell{1}{1} = {r=2}{},
  cell{1}{2} = {c=8}{},
  cell{1}{10} = {r=2}{},
  cell{1}{11} = {r=2}{},
  cell{2}{2} = {c},
  cell{2}{3} = {c},
  cell{2}{4} = {c},
  cell{2}{5} = {c},
  cell{2}{6} = {c},
  cell{2}{8} = {c},
  cell{2}{9} = {c},
  cell{3}{1} = {c},
  cell{3}{2} = {c},
  cell{3}{3} = {c},
  cell{3}{4} = {c},
  cell{3}{5} = {c},
  cell{3}{6} = {c},
  cell{3}{8} = {c},
  cell{3}{9} = {c},
  cell{3}{10} = {c},
  cell{4}{1} = {c},
  cell{4}{2} = {c},
  cell{4}{3} = {c},
  cell{4}{4} = {c},
  cell{4}{5} = {c},
  cell{4}{6} = {c},
  cell{4}{8} = {c},
  cell{4}{9} = {c},
  cell{4}{10} = {c},
  cell{5}{1} = {c},
  cell{5}{2} = {c},
  cell{5}{3} = {c},
  cell{5}{4} = {c},
  cell{5}{5} = {c},
  cell{5}{6} = {c},
  cell{5}{8} = {c},
  cell{5}{9} = {c},
  cell{5}{10} = {c},
  cell{6}{1} = {c},
  cell{6}{10} = {c},
  hline{1,6} = {-}{0.08em},
  hline{3} = {-}{},
}
\# & \textbf{Session-wise Harmonic Mean~(\%)~↑ } &               &               &               &               &               &               &               &        \ahm{} & \acc{}      \\
                & 1                                           & 2             & 3             & 4             & 5             & 6             & 7             & 8             & &              \\
0               & 69.3                                        & 47.9          & 42.3          & 34.9          & 31.2          & 28.0          & 24.8          & 24.4          & 37.8         & 54.6 \\
1               & 64.4                                        & 53.4          & 48.7          & 48.9          & 45.4          & 40.5          & 40.7          & 43.5          & 48.2         & 64.9 \\
5               & \textbf{68.7}                               & \textbf{63.9} & \textbf{60.9} & \textbf{58.0} & \textbf{55.3} & \textbf{52.4} & \textbf{52.7} & \textbf{53.1} & \textbf{58.1} & \textbf{67.1}
\end{tblr}
}}}
\caption{Number of saved exemplars (\#) for incremental sessions.}
\label{tab:exemplarablation}
\end{table}

\section{More Results}
\label{sec:more_results}
\myparagraph{Base and incremental accuracy breakdown.}
We show our SOTA results with a base and incremental session accuracy breakdown for all sessions in tables \labelcref{tab:minet_baseincbreakdown,tab:cifar_baseincbreakdown,tab:cub_baseincbreakdown}. Note that, for Imprinted Weights (IW)~\cite{qi2018low} we use the implementation of a decoupled learning strategy from \cite{de2021continual} and we initialise the method with our model pretrained during Phase 1.
We note that LIMIT~\cite{zhou2022few} has been unintentionally left out from figure 4 (main) for CUB200.
Furthermore, we report the average accuracy metric to provide a comprehensive overview of our results. Note that the considerable data imbalance between base and incremental classes has an impact on this metric. The improved accuracy especially in the base classes, as illustrated in the breakdown tables, contributes to the overall enhancement of this measure.

\myparagraph{Confusion matrices.}
In figure~\ref{fig:sessconfusionmatrix}, we compare session-wise confusion matrices for a) \method, b) NC-FSCIL~\cite{Yang2023Neural}, and c) BiDist~\cite{zhao2023few} during the final session of mini-ImageNet. Our benchmark involves assessing \method against its two closest competitors. \method plays a crucial role in finding a delicate balance between preserving knowledge of base classes and efficiently learning new ones, showcasing significantly enhanced learning capabilities in incremental classes. Notably, other methods exhibit a strong bias towards the base classes due to low transferability, while our pretraining session establishes a robust backbone. Moreover, our space reservation scheme, along with strong separation using perturbed targets and robust contrastive learning, enables us to learn a highly performant learner.

\section{Theory of Orthogonality}
\label{sec:ortho}

This section covers the mathematical theory of orthogonality of independent vectors in high dimensional space.
Let $X_1, X_2, \ldots, X_n$ be independent and identically distributed (i.i.d.) random vectors sampled from a normal distribution with mean $0$ and variance $1$. These vectors are in $\mathbb{R}^n$.
The claim is that these vectors are mutually orthogonal on the unit sphere. To prove this, let's first establish that the vectors are normalized to have a length of $1$.

Given a vector $X_i = (X_{i1}, X_{i2}, \ldots, X_{in})$, its length is given by:
\[
\|X_i\| = \sqrt{X_{i1}^2 + X_{i2}^2 + \ldots + X_{in}^2}
\]
Since each component $X_{ij}$ is independently sampled from a normal distribution with mean $0$ and variance $1$, the expected value of $X_{ij}^2$ is $1$. Therefore, the expected value of the length squared is:
\[
\mathbb{E}[\|X_i\|^2] = \mathbb{E}[X_{i1}^2 + X_{i2}^2 + \ldots + X_{in}^2] = n
\]
This means that $\frac{1}{\sqrt{n}} X_i$ has a length of $1$ in expectation. Now, let's consider the inner product of two different vectors $X_i$ and $X_j$ (where $i \neq j$):
\[
\mathbb{E}[X_i \cdot X_j] = \sum_{k=1}^{n}\sum_{k'=1}^{n} \mathbb{E}[X_{ik} \cdot X_{jk'}]
\]

Since $X_{ik}$ and $X_{jk}$ are independent for $i \neq j$, the cross-terms in the summation will have an expected value of $0$, and the only non-zero terms will be the ones where $k = k'$, resulting in:
\[
\mathbb{E}[X_i \cdot X_j] = \sum_{k=1}^{n} \mathbb{E}[X_{ik} \cdot X_{jk}] = \sum_{k=1}^{n} \mathbb{E}[X_{ik}^2] \delta_{ij} = \delta_{ij} \cdot n
\]
where $\delta_{ij}$ is the Kronecker delta. Therefore, the expected value of the inner product is $n$ if $i = j$ and $0$ otherwise. This means that the vectors are orthogonal in expectation. It is important to note that this property specifically holds for vectors drawn from a normal distribution with mean 0 and variance 1.

\begin{figure}
\includegraphics[width=0.5\textwidth]{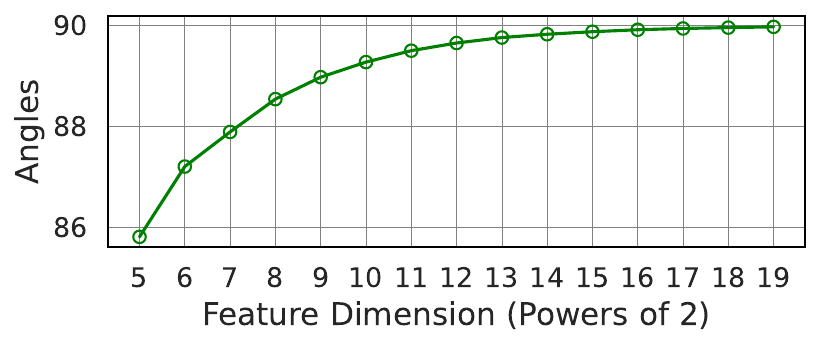} 
\caption{Practical effect of dimensions on average pair-wise angle of a 100 independant random vectors.}
\label{fig:angleagainstdim}
\end{figure}

\begin{figure*}[t]
  \begin{subfigure}{0.3\textwidth}
    \centering
    \includegraphics[width=\linewidth]{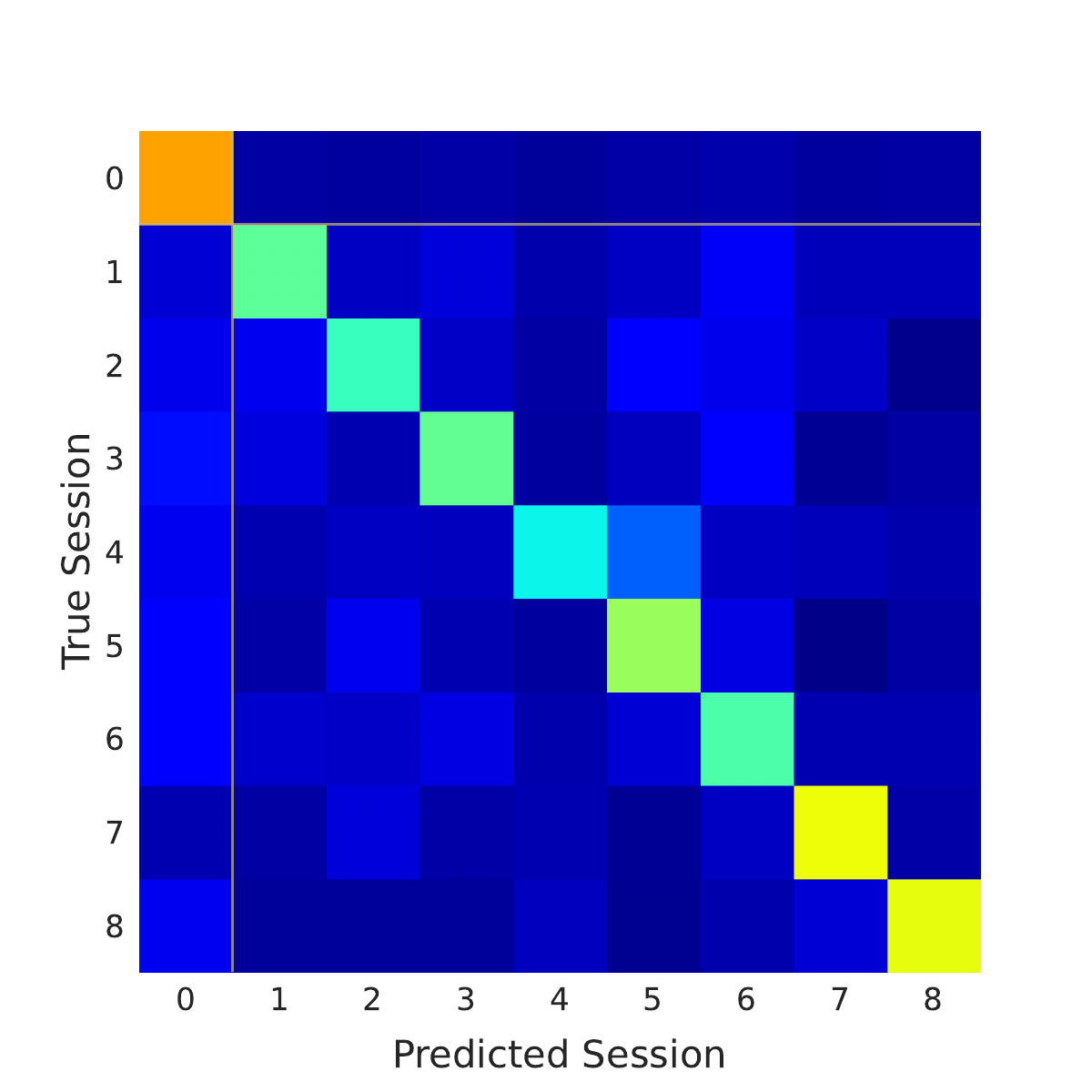}
    \caption{OrCo}
  \end{subfigure}
  \hfill
  \begin{subfigure}{0.3\textwidth}
    \centering
    \includegraphics[width=\linewidth]{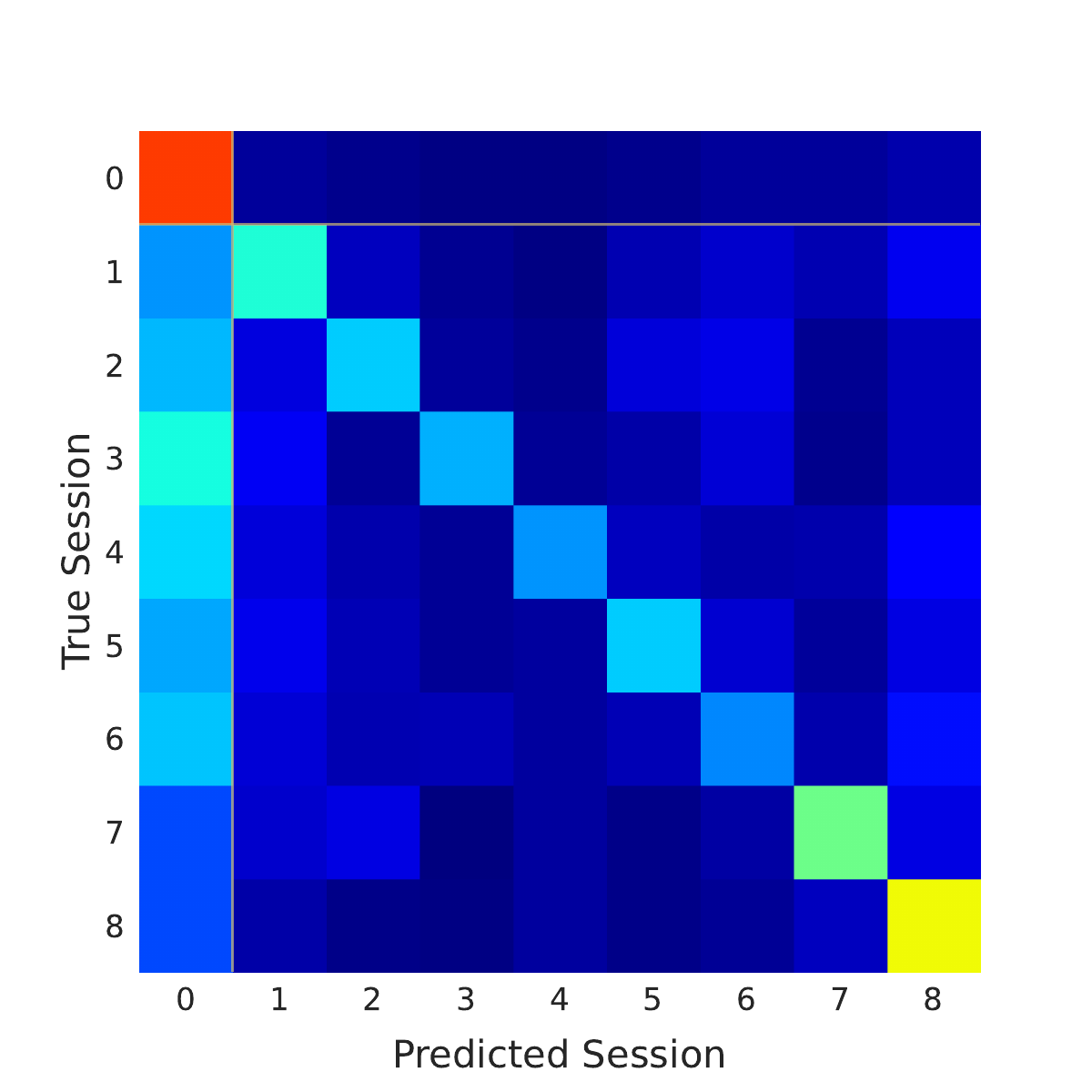}
    \caption{NC-FSCIL}
  \end{subfigure}
  \hfill
  \begin{subfigure}{0.375\textwidth}
    \centering
    \includegraphics[width=\linewidth]{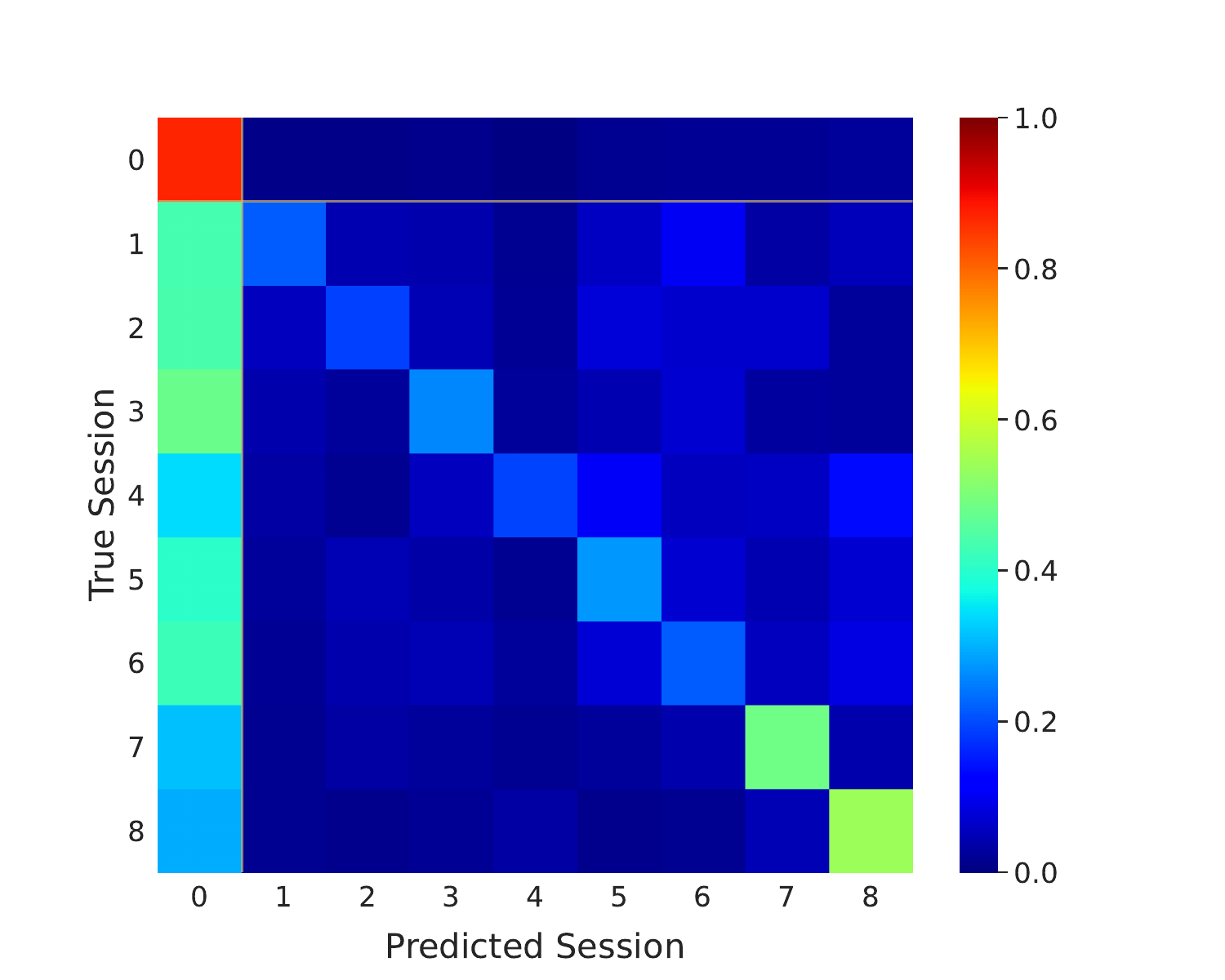}
    \caption{BiDist}
  \end{subfigure}
\caption{Visualising the session-wise confusion matrix for mini-ImageNet using a) OrCo, b) NC-FSCIL~\cite{Yang2023Neural}, and c) BiDist~\cite{zhao2023few}. Each matrix demonstrates the predictive accuracy for base and incremental sessions, separated by yellow lines. High values on the diagonal (indicative of correct session predictions) are contrasted with low off-diagonal values (representing misclassifications). The first column in each matrix highlights potential prediction bias towards base classes. Our method's performance, as illustrated, demonstrates both high diagonal accuracy and a balanced approach in reducing base class bias, as compared to the results of the competing methods.}
\label{fig:sessconfusionmatrix}
\end{figure*}

The figure \ref{fig:angleagainstdim} shows the practical effects of the above theory which yields near orthogonality but not perfect orthogonality. Only near feature dimension = $2^{15}$ do we generate nearly orthogonal vectors. Which would lead to the projection head having $\sim{68}$ million parameter making our framework incomparable to other methods. Explicit orthogonality as we have shown previously in table 3 (main) yields better results which is why we explore this practical constraint.

\section{Contrastive losses}
\label{sec:contrastive_losses}

\myparagraph{Supervised Contrastive Loss.}
Given a set of sample-label pairs $(x_i, y_i) \in Z^{SCL}$, we define the positive set $P_i^{SCL}$ for $x_i$ as the collection of pairs $(x_j, y_j)$ where $j$ varies over all instances such that $y_j = y_i$. Correspondingly, the negative set $N_i^{SCL}$ is defined as $Z^{SCL} \setminus P_i^{SCL}$. Then, supervised contrastive loss (SCL) is defined as:

\begin{equation*}
    \mathcal{L}_{SCL}(i; \theta) = \frac{-1}{|P_i^{SCL}|} \sum_{x_j \in P_i^{SCL}} \text{log} \frac{\text{exp}(x_i\cdot x_j / \tau)}{\sum\limits_{x_k \in N_i^{SCL}} \text{exp}(x_i \cdot x_k / \tau) } .
\label{eq:supconloss_supmat}
\end{equation*}

\myparagraph{Self-Supervised Contrastive Loss.}
Given a set of samples $x_i \in Z^{SSCL}$, we define a positive for $x_i$ as $A(x_i)$ where $A(\cdot)$ is a random transformation. Then, self-supervised contrastive loss (SSCL) loss is defined as:
\begin{equation*}
    \mathcal{L}_{SSCL}(i; \theta) = -1 \cdot \text{log} \frac{\text{exp}(x_i\cdot A(x_i) / \tau)}{\sum\limits_{x_k \in Z^{SSCL}, i \neq k} \text{exp}(x_i \cdot x_k / \tau) } .
\end{equation*}

\section{Orthogonality Loss}
\label{sec:detail_orth_loss}

In this section, we further expand on the orthogonality loss in our framework. We employ the orthogonality loss as an implicit geometric constraint on the set $O$ predicated on the current batch. $O$ contains the following: mean features for all classes within batch $\mu_j$, assigned targets not represented within batch and all unassigned targets $T^i_u$. 

Formally, let us assume the session $i$ with data $D^i$ and classes $C^i$. In order to define a set $O$ we compute some preliminaries. Firstly, for every training batch $B$ we compute the within-batch mean for all data features. This is computed as:

\begin{equation}
    \mu_j = \frac{1}{|C^j|}\sum^{|C^j|}_{k=0}z_k, \space \forall j \in C^i_{B}
\end{equation}

where $C^i_{B} \in C^i$ refers to all classes appearing in this particular batch. The combined set of all means can then be termed $M_{B}$. For the classes that did not appear in this batch we define as $\neg{C^i_B} = C^i \setminus C^i_B$. Subsequently we define a mapping function from seen class labels to the assigned pseudo target.

\begin{equation}
    h : C^i \rightarrow T^i
\end{equation}

We incorporate the remaining real data by adding the following set of assigned pseudo targets as $\neg{T}^i_B = h(\neg{C^i_B})$. For completeness, we combine the above with the unassigned targets $T^i_u$ leading to the following definition of $O$:

\begin{equation}
    O = \{M_B \cup \neg{T}^i_B \cup T^i_u \mid B\}
\end{equation}

Finally, the orthogonality loss takes the form:

\begin{equation}
    \mathcal{L}_{ORTH}(O) = \frac{1}{|O|}\sum^{|O|}_{i=1} \text{log} \sum^{|O|}_{j=1} e^{o_i \cdot o_j / \tau_o}
    \label{eq:lorth}, o_i, o_j \in O
\end{equation}

In essence, the orthogonality loss introduces a subtle geometric constraint between real class features and the pseudo-targets. Additionally the batch-wise construction helps regularise the loss function.

Although the improvement from the Orthogonality loss are not prominent like in PSCL, it remains measurable and consistent across settings (e.g. in table \ref{tab:lossablation} line 1 vs line 2, line 3 vs line 4), and thus contributes to our results. Additionally, we would like to highlight that each loss term in equation \ref{eq:uniconloss} incorporates orthogonality constraints either implicitly or explicitly. E.g. the orthogonality enforced on the pseudo-targets implies an implicit orthogonality among the features as they are pushed to these targets.

\section{Further Implementation Details}
\label{sec:impl_details}
\myparagraph{Model parameters.}
Our representation learning framework is composed of:

\begin{itemize}

\item Encoder/Backbone Network ($f$) We use the ResNet18 and 12 \cite{he2016deep} variant for our experiments. Depending on the variant of the encoder, the representation vector has output dimensions $D_E$ = 512 or 640, respectively. In table \ref{tab:architectures} we compare our choice of architecture against other FSCIL studies.  

\item  Projection Network (projector)($g$). Following the encoder, a projection MLP maps the representation vector from $f$ to the contrastive subspace. We use a two layer projection head with a hidden dimension of size 2048. By convention output projections are normalised to the hypersphere. For simplicity, the dimension of the projected hyper sphere is initialised as $d=2\lceil{log(C)}\rceil$ where $C$ is the total number of classes for our FSCIL task. Following convention, we assume normalised feature vectors.
\end{itemize}

\myparagraph{Training details.} 
For CUB200 dataset, we skip the pretraining phase given that it is common in literature to use pretrained ImageNet weights for the backbone \cite{de2021continual, Yang2023Neural, zhao2023few, Zhou2022}. For any incremental sessions ($>0$) we finetune only the projection head from phase 2. For the PSCL loss we choose a perturbation magnitude $\lambda_{pert} = \text{1e-2}$ for our experiments and perturb only the incremental targets and unassigned targets.
We train the projection head for 10 epochs in the 0-th incremental session and 100 epochs for all following sessions. Cosine scheduling is employed with warmup for a few epochs for all our phases with a maximum learning rate set to $0.4$ in phase 1, $0.25$ in phase 2 and $0.1$ in phase 3. 
Given the equation 4 (main),
we double our batch size by over sampling target perturbations such that the number of perturbed targets is always the same as the original training batch size. Orthogonality loss is applied batch wise. More concretely, the orthogonality loss takes as input, the within class average features inside a batch along with any other pseudo targets not in the batch. 
Cross entropy is applied exclusively to incremental class data as base classes are already aligned. 

For pseudo target generation we employ an SGD optimiser, with 1e-2 learning rate for 2000 epochs to minimize the loss. 
For CUB200 the dimension of the projection head is higher which constitutes longer training cycle to fully orthogonalize the pseudo targets. CUB200 was most susceptible to forgetting for which reason we ensured that base classes were incorporated inside the CE loss component of the loss function. With the lack of a pretrain step in CUB200, we must also finetune the backbone during phase 2 to align base classes while also capturing specific representation which is important for learning effectively on a fine-grained dataset.
For pretraining, we use RandAug~\cite{cubuk2020randaugment} for mini-Imagenet and AutoAugment policy~\cite{cubuk2018autoaugment} for CIFAR100. 
Additionally, following the implementation of \cite{Zhou2022} we apply auto augment policy for CIFAR100 during the incremental sessions as well.

\begin{table}
\centering
\small
\begin{tabular}{l c c c}
\toprule
\multirow{2}{*}{Method} & \multicolumn{3}{c}{Model} \\
\cline{2-4}
 & mini-ImageNet & CIFAR100 & CUB200 \\
\midrule
IW~\cite{qi2018low} & ResNet18 & ResNet12 & ResNet18 \\
FACT~\cite{Zhou2022} & ResNet18 & ResNet20 & ResNet18 \\
CEC~\cite{Zhang2021} & ResNet18 & ResNet20 & ResNet18 \\
C-FSCIL~\cite{hersche2022cfscil} & ResNet12 & ResNet12 & - \\
LIMIT~\cite{zhou2022few} & ResNet18 & ResNet20 & ResNet18 \\
LCwoF~\cite{Anna} & ResNet18 & - & - \\
BiDist~\cite{zhao2023few} & ResNet18 & ResNet18 & ResNet18 \\
NC-FSCIL~\cite{Yang2023Neural} & ResNet12 & ResNet12 & ResNet18 \\
OrCo & ResNet18 & ResNet12 & ResNet18 \\
\bottomrule

\end{tabular}
\caption{ResNet architectures used in FSCIL literature}
\label{tab:architectures}
\end{table}

\begin{table}[h]
\centering
\small
\begin{tabular}{c c c c c}
\toprule
\multirow{2}{*}{Architecture} & \multirow{2}{*}{Parameter Count} & \multirow{2}{*}{aHM} & \multirow{2}{*}{Base Acc} \\
& (million) & & \\
\midrule
ResNet-18 & \textbf{12.49} & 58.12 & 83.30 \\
ResNet-12 & 15.06 & \textbf{59.30} & \textbf{83.65} \\
\bottomrule
\end{tabular}
\caption{Comparing ResNet-12 to ResNet-18}
\label{tab:12vs18}
\end{table}

\section{ResNet 12 with Mini-ImageNet}
In this section we measure the efficacy of our method on a different backbone. More specifically we train a ResNet-12 backbone used by \cite{Yang2023Neural} with our method. In table \ref{tab:12vs18} we show the results. Our reported results with ResNet-18 are 58.12, while the results with ResNet-12 are 59.30 indicating an improvement with the wider ResNet-12 architecture. ResNet-18 remains as our elected architecture due to its prominence in prior works, low parameter count, while still maintaining state-of-the art performance.

\begin{table*}
\centering
\small
\begin{tabular}{l l c c c c c c c c c l c}
\toprule
\multirow{2}{*}{Method} & \multirow{2}{*}{Class Group} &  & \multicolumn{8}{c}{\textbf{Session-wise Accuracy (\%)}} & \multirow{2}{*}{\textbf{Means}}  & \multirow{2}{*}{{aACC}} \\
\cline{3-11}
 &  & \textbf{0} & \textbf{1} & \textbf{2} & \textbf{3} & \textbf{4} & \textbf{5} & \textbf{6} & \textbf{7} & \textbf{8} & &  \\
\midrule

\multirow{2}{*}{IW~\cite{qi2018low}} & Base & \textbf{83.10} & \textbf{81.17} & \textbf{80.58} & \textbf{79.93} & \textbf{79.55} & \textbf{78.88} & \textbf{78.38} & \textbf{78.08} & \textbf{77.25} & \textbf{79.66}  & \multirow{2}{*}{{68.77}}\\

 & Incremental & - & 35.60 & 31.30 & 32.27 & 32.65 & 31.16 & 29.13 & 29.91 & 32.40 & 31.80 & \\
\hdashline[1pt/1pt]
\multirow{2}{*}{FACT~\cite{Zhou2022}} & Base & 75.78 & 75.22 & 74.83 & 74.47 & 74.30 & 74.05 & 73.82 & 73.47 & 73.37 & 74.37 & \multirow{2}{*}{{60.86}} \\

 & Incremental & - & 16.60 & 17.10 & 17.20 & 15.15 & 13.24 & 11.93 & 12.17 & 12.43 & 14.48 & \\
\hdashline[1pt/1pt]
\multirow{2}{*}{CEC~\cite{Zhang2021}} & Base & 72.17 & 70.77 & 70.05 & 69.53 & 69.27 & 68.95 & 68.65 & 68.25 & 67.95 & 69.51 & \multirow{2}{*}{{59.57}} \\

 & Incremental & - & 20.60 & 20.60 & 19.93 & 19.75 & 17.68 & 16.63 & 16.80 & 17.18 & 18.65 &  \\
\hdashline[1pt/1pt]
\multirow{2}{*}{C-FSCIL~\cite{hersche2022cfscil}} & Base & 76.60 & 76.15 & 74.70 & 73.82 & 73.18 & 71.10 & 70.08 & 68.27 & 67.58 & 72.39 & \multirow{2}{*}{{61.21}}\\

 & Incremental & - & 5.20 & 11.90 & 17.80 & 20.40 & 23.08 & 23.37 & 26.06 & 26.35 & 19.27 & \\
\hdashline[1pt/1pt]
\multirow{2}{*}{LIMIT~\cite{zhou2022few}} & Base & 73.27 & 70.43 & 69.47 & 68.68 & 68.18 & 67.73 & 67.30 & 67.07 & 66.68 & 68.76  & \multirow{2}{*}{{58.04}}\\

 & Incremental & - & 29.80 & 24.00 & 22.73 & 21.85 & 19.72 & 19.40 & 19.71 & 20.18 & 22.17 & \\
\hdashline[1pt/1pt]
\multirow{2}{*}{LCwoF~\cite{Anna}} & Base & 64.45 & 57.33 & 53.31 & 52.87 & 51.38 & 48.25 & 47.60 & 47.51 & 47.73 & 52.27 & \multirow{2}{*}{{46.80}} \\

 & Incremental & - & 32.20 & 30.70 & 31.00 & 31.12 & 29.68 & 28.27 & 27.34 & 27.65 & 29.75 & \\
\hdashline[1pt/1pt]
\multirow{2}{*}{BiDist~\cite{zhao2023few}} & Base & 74.67 & 73.63 & 72.50 & 71.03 & 70.63 & 70.37 & 68.70 & 67.98 & 69.25 & 70.97 & \multirow{2}{*}{{61.25}} \\

 & Incremental & - & 32.60 & 31.40 & 30.33 & 30.30 & 25.48 & 25.23 & 27.09 & 25.10 & 28.44 & \\
\hdashline[1pt/1pt]
\multirow{2}{*}{NC-FSCIL~\cite{Yang2023Neural}} & Base & 84.37 & 78.25 & 76.00 & 75.73 & 74.80 & 75.42 & 75.52 & 75.13 & 74.77 & 76.67 & \multirow{2}{*}{{67.82}} \\

 & Incremental & - & 51.80 & 51.00 & 44.33 & 41.20 & 37.04 & 34.20 & 33.37 & 32.77 & 40.71 & \\
\hdashline[1pt/1pt]
\multirow{2}{*}{OrCo} & Base & 83.30 & 76.40 & 74.10 & 72.00 & 71.20 & 70.50 & 69.20 & 68.10 & 67.60 & 72.49 & \multirow{2}{*}{{67.14}} \\

 & Incremental & \textbf{-} & \textbf{62.40} & \textbf{56.10} & \textbf{52.80} & \textbf{48.90} & \textbf{45.40} & \textbf{42.20} & \textbf{42.90} & \textbf{43.80} & \textbf{49.31} & \\
\bottomrule
\end{tabular}
\caption{Base and Incremental accuracy shown per session for mini-ImageNet.}
\label{tab:minet_baseincbreakdown}
\end{table*}

\begin{table*}
\centering
\small
\begin{tabular}{l  l  c  c  c  c  c  c  c  c  c  l c }
\toprule
\multirow{2}{*}{Method} & \multirow{2}{*}{Class Group} &  & \multicolumn{8}{c}{\textbf{Session-wise Accuracy (\%)}} & \multirow{2}{*}{\textbf{Means}} & \multirow{2}{*}{{aACC}} \\
\cline{3-11}
 &  & \textbf{0} & \textbf{1} & \textbf{2} & \textbf{3} & \textbf{4} & \textbf{5} & \textbf{6} & \textbf{7} & \textbf{8} & &  \\
\midrule

\multirow{2}{*}{IW~\cite{qi2018low}} & Base & 78.58 & 75.45 & 75.15 & 74.65 & 74.38 & 74.07 & 73.80 & 73.55 & 73.03 & 74.74 & \multirow{2}{*}{{64.05}} \\

 & Incremental & - & 29.40 & 31.90 & 28.53 & 27.60 & 26.36 & 27.20 & 26.91 & 25.88 & 27.97 & \\
\hdashline[1pt/1pt]
\multirow{2}{*}{C-FSCIL~\cite{hersche2022cfscil}} & Base & 77.35 & 76.70 & 76.17 & 75.52 & 75.35 & 74.22 & 73.92 & 73.63 & 72.87 & 75.08 & \multirow{2}{*}{{61.42}} \\

 & Incremental & - & 19.80 & 17.40 & 16.20 & 13.65 & 14.92 & 14.53 & 13.66 & 14.00 & 15.52 \\
\hdashline[1pt/1pt]
\multirow{2}{*}{LIMIT~\cite{zhou2022few}} & Base & 79.63 & 75.40 & 74.47 & 73.70 & 73.22 & 72.52 & 72.22 & 72.02 & 71.32 & 73.83 & \multirow{2}{*}{{61.66}}\\

 & Incremental & - & 27.20 & 24.60 & 21.47 & 21.00 & 20.76 & 21.10 & 20.54 & 20.13 & 22.10 \\
\hdashline[1pt/1pt]
\multirow{2}{*}{CEC~\cite{Zhang2021}} & Base & 72.93 & 72.13 & 71.42 & 70.72 & 70.12 & 69.20 & 68.67 & 68.43 & 67.75 & 70.15 & \multirow{2}{*}{{59.53}}\\

 & Incremental & - & 29.60 & 27.00 & 22.60 & 21.80 & 22.40 & 22.13 & 21.66 & 21.08 & 23.53 \\
\hdashline[1pt/1pt]
\multirow{2}{*}{FACT~\cite{Zhou2022}} & Base & 78.72 & 76.23 & 75.30 & 74.63 & 73.90 & 73.07 & 72.58 & 72.28 & 71.73 & 74.27 & \multirow{2}{*}{{62.55}} \\

 & Incremental & - & 29.80 & 25.60 & 21.20 & 20.70 & 20.24 & 22.33 & 21.69 & 21.95 & 22.94 \\
\hdashline[1pt/1pt]
\multirow{2}{*}{BiDist~\cite{zhao2023few}} & Base & 69.68 & 68.45 & 67.55 & 66.47 & 65.80 & 64.87 & 64.77 & 64.27 & 64.50 & 66.26 & \multirow{2}{*}{{56.91}} \\

 & Incremental & - & 36.80 & 31.20 & 28.80 & 25.75 & 24.36 & 22.87 & 22.43 & 20.35 & 26.57 \\
\hdashline[1pt/1pt]
\multirow{2}{*}{NC-FSCIL~\cite{Yang2023Neural}} & Base & \textbf{82.52} & \textbf{79.55} & \textbf{78.63} & \textbf{77.98} & \textbf{77.60} & \textbf{75.98} & \textbf{74.45} & \textbf{75.18} & \textbf{73.98} & \textbf{77.32} & \multirow{2}{*}{{67.50}} \\

 & Incremental & - & 44.00 & 41.60 & 36.47 & 31.95 & 31.32 & 33.97 & 31.31 & 29.30 & 34.99 \\
\hdashline[1pt/1pt]
\multirow{2}{*}{OrCo} & Base & 80.08 & 67.37 & 68.12 & 63.30 & 63.40 & 63.93 & 61.45 & 61.08 & 58.22 & 65.22 & \multirow{2}{*}{{62.11}} \\
 & Incremental & - & \textbf{77.60} & \textbf{60.20} & \textbf{51.67} & \textbf{48.90} & \textbf{45.80} & \textbf{48.23} & \textbf{44.94} & \textbf{43.15} & \textbf{52.56} \\
\bottomrule
\end{tabular}
\caption{Base and Incremental accuracy shown per session for CIFAR100.}
\label{tab:cifar_baseincbreakdown}
\end{table*}

\begin{table*}
\centering
\scalebox{1}{ 
\centering
\setlength{\tabcolsep}{1mm}{
\resizebox{0.98\linewidth}{!}{
\begin{tabular}{ l  l  c  c  c  c  c  c  c  c  c  c  c  l c }
\toprule
\multirow{2}{*}{Method} & \multirow{2}{*}{Class Group} & \multicolumn{11}{c}{\textbf{Session-wise Accuracy (\%)}} & \multirow{2}{*}{\textbf{Means}} & \multirow{2}{*}{{aACC}} \\
\cline{3-13}
 &  & \textbf{0} & \textbf{1} & \textbf{2} & \textbf{3} & \textbf{4} & \textbf{5} & \textbf{6} & \textbf{7} & \textbf{8} & \textbf{9} & \textbf{10} &  \\
\midrule
\multirow{2}{*}{IW~\cite{qi2018low}} & Base & 67.53 & 67.07 & 66.83 & 66.55 & 66.48 & 66.31 & 66.13 & 65.99 & 65.85 & 65.85 & 65.75 & 66.39 & \multirow{2}{*}{{59.72}} \\

 & Incremental & - & 29.03 & 27.74 & 25.00 & 25.95 & 26.61 & 26.51 & 25.45 & 24.34 & 26.08 & 26.93 & 26.36 \\
\hdashline[1pt/1pt]
\multirow{2}{*}{CEC~\cite{Zhang2021}} & Base & 75.64 & 74.27 & 73.88 & 73.64 & 72.66 & 72.31 & 71.75 & 71.09 & 70.98 & 70.64 & 70.46 & 72.48 & \multirow{2}{*}{{61.33}} \\

 & Incremental & - & 45.16 & 41.52 & 33.1 & 36.34 & 33.1  & 34.17 & 34.34 & 32.96 & 34.41 & 34.16 & 35.93 \\
\hdashline[1pt/1pt]
\multirow{2}{*}{BiDist~\cite{zhao2023few}} & Base & 75.98 & 74.23 & 73.71 & 73.85 & 73.08 & 72.35 & 71.68 & 71.65 & 71.68 & 71.12 & 70.36 & 72.70 & \multirow{2}{*}{{62.91}} \\

 & Incremental & - & 55.20 & 46.11 & 38.19 & 41.67 & 37.62 & 38.11 & 39.28 & 36.81 & 39.73 & 39.83 & 41.26 \\
\hdashline[1pt/1pt]
\multirow{2}{*}{FACT~\cite{Zhou2022}} & Base & 77.23 & 75.04 & 74.83 & 74.79 & 74.41 & 74.20 & 73.60 & 73.46 & 73.22 & 72.87 & 72.84 & 74.22 & \multirow{2}{*}{{64.42}} \\

 & Incremental & - & 53.05 & 47.17 & 38.08 & 40.21 & 38.37 & 39.66 & 40.25 & 38.30 & 40.11 & 39.93 & 41.51 \\
\hdashline[1pt/1pt]
\multirow{2}{*}{LIMIT~\cite{zhou2022few}} & Base & 79.63 & \textbf{79.02} & \textbf{78.81} & \textbf{78.77} & \textbf{78.39} & \textbf{78.04} & \textbf{77.72} & \textbf{77.51} & \textbf{77.24} & \textbf{76.68} & 73.46 & \textbf{77.75} & \multirow{2}{*}{{65.49}}\\

 & Incremental & - & 49.82 & 44.88 & 37.38 & 39.35 & 37.35 & 38.40 & 40.01 & 38.47 & 39.92 & 42.15 & 40.77 \\
\hdashline[1pt/1pt]
\multirow{2}{*}{NC-FSCIL~\cite{Yang2023Neural}} & Base & \textbf{80.45} & 76.89 & 77.62 & 78.63 & 77.23 & 77.13 & 76.85 & 76.29 & 76.61 & 75.94 & \textbf{76.19} & 77.26 & \multirow{2}{*}{{67.29}} \\

 & Incremental & - & 66.67 & 45.41 & 42.59 & 45.88 & 41.72 & 44.11 & 44.99 & 41.16 & 42.66 & 43.07 & 45.83 \\
\hdashline[1pt/1pt]
\multirow{2}{*}{OrCo} & Base & 75.59 & 65.54 & 63.79 & 66.76 & 64.91 & 65.05 & 65.50 & 65.01 & 66.24 & 66.27 & 66.62 & 66.48  & \multirow{2}{*}{{62.36}} \\

 & Incremental & - & \textbf{79.93} & \textbf{65.37} & \textbf{53.47} & \textbf{55.41} & \textbf{51.03} & \textbf{51.31} & \textbf{50.90} & \textbf{47.69} & \textbf{49.70} & \textbf{49.25} & \textbf{55.41} \\
\bottomrule
\end{tabular}
}}}
\caption{Base and Incremental accuracy shown per session for CUB200.}
\label{tab:cub_baseincbreakdown}
\end{table*}

\end{document}